\documentclass[runningheads]{llncs}

\usepackage[mobile]{accv}

\usepackage{accvabbrv}
\usepackage{multirow}

\usepackage{graphicx}
\usepackage{booktabs}

\usepackage[accsupp]{axessibility}  
\usepackage{float}
\usepackage[symbol]{footmisc}

\usepackage[pagebackref,breaklinks,colorlinks,citecolor=accvblue]{hyperref}

\usepackage{orcidlink}

\begin{document}

\title{PrimeDepth: Efficient Monocular Depth Estimation with a Stable Diffusion Preimage} 

\titlerunning{PrimeDepth}

\author{Denis Zavadski${}^\ast$\inst{} \and
Damjan Kalšan${}^\ast$\inst{}\and
Carsten Rother\inst{}}

\authorrunning{D.Zavadski et al.}

\institute{Computer Vision and Learning Lab, IWR, Heidelberg University, Germany\\
\email{\{name.surname\}@iwr.uni-heidelberg.de}}

\maketitle
\footnotetext[1]{Equal Contribution}
\renewcommand*{\thefootnote}{\arabic{footnote}}

\begin{abstract}

This work addresses the task of zero-shot monocular depth estimation. A recent advance in this field has been the idea of utilising Text-to-Image foundation models, such as Stable Diffusion~\cite{LDM_Rombach_2022_CVPR}. 
Foundation models provide a rich and generic image representation, and therefore, little training data is required to reformulate them as a depth estimation model that predicts highly-detailed depth maps and has good generalisation capabilities. However, the realisation of this idea has so far led to approaches which are, unfortunately, highly inefficient at test-time due to the underlying iterative denoising process. In this work, we propose a different realisation of this idea and present PrimeDepth, a method that is highly efficient at test time while keeping, or even enhancing, the positive aspects of diffusion-based approaches. Our key idea is to extract from Stable Diffusion a rich, but frozen, image representation by running a single denoising step. This representation, we term preimage, is then fed into a refiner network with an architectural inductive bias, before entering the downstream task. We validate experimentally that PrimeDepth is two orders of magnitude faster than the leading diffusion-based method, Marigold~\cite{Marigold_Ke_2024_CVPR}, while being more robust for challenging scenarios and quantitatively marginally superior. 
Thereby, we reduce the gap to the currently leading data-driven approach, Depth Anything~\cite{Yang2024_DepthAnything}, which is still quantitatively superior, but predicts less detailed depth maps and requires 20 times more labelled data. Due to the complementary nature of our approach, even a simple averaging between PrimeDepth and Depth Anything predictions can improve upon both methods and sets a new state-of-the-art in zero-shot monocular depth estimation. In future, data-driven approaches may also benefit from integrating our preimage. 

  \keywords{Monocular Depth Estimation \and Diffusion Models}
\end{abstract}

\begin{figure}[tb!]
  \centering
  \includegraphics[width=.98\linewidth]{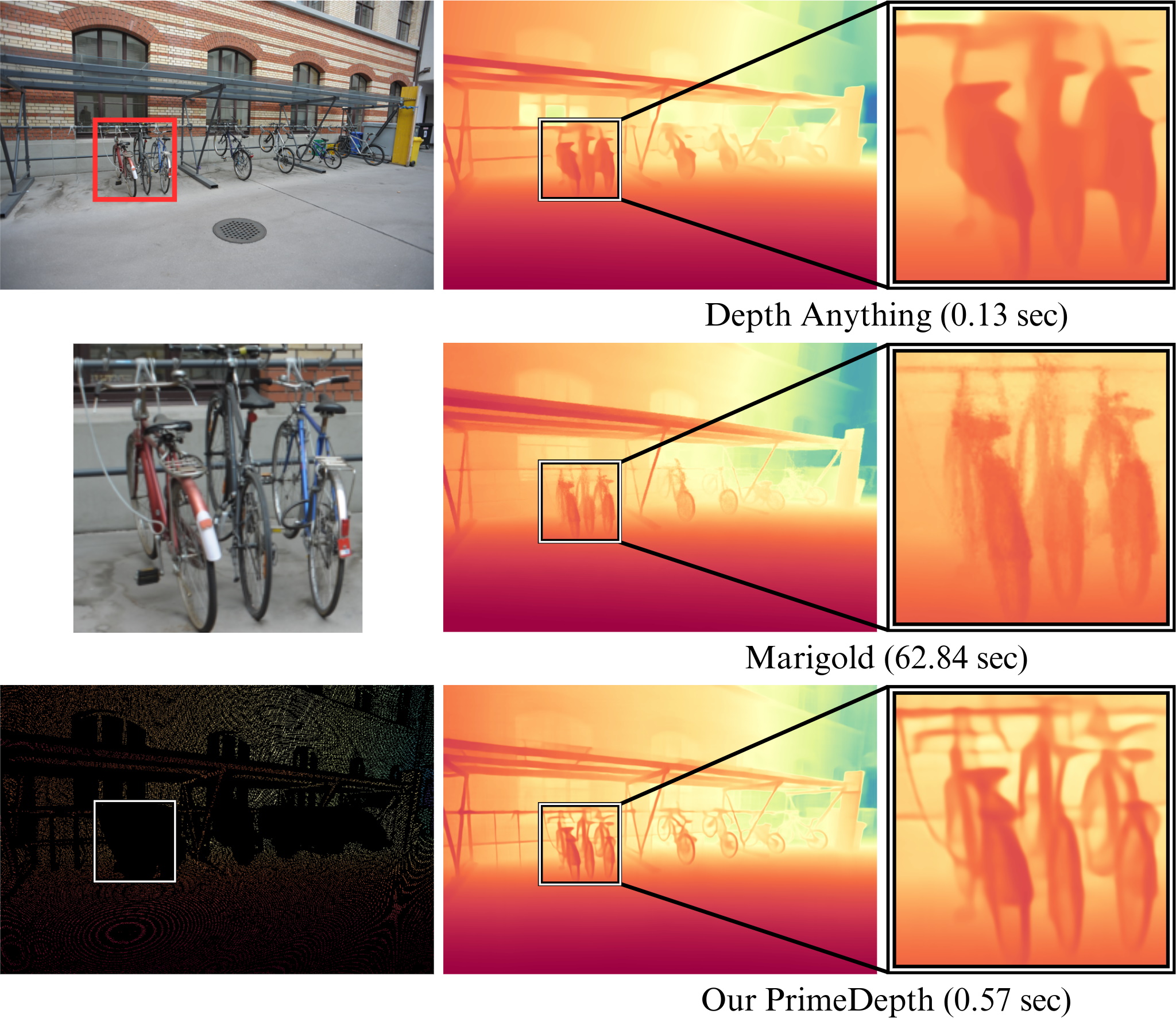}
  \caption{
  Results of Depth Anything~\cite{Yang2024_DepthAnything} (top row), Marigold~\cite{Marigold_Ke_2024_CVPR} (middle row) and our PrimeDepth (bottom row), for a challenging scene from the ETH3D Dataset~\cite{schops2017ETH3D}. While our non-optimised method is fast at test time (0.57 sec\protect\footnotemark), as well as Depth Anything (0.13 sec), Marigold is rather slow (62.84 sec). Runtimes were measured on an A100 GPU.  
  Visually, our result shows most details, i.e. more detailed than Depth Anything and less grainy than Marigold, which however is not reflected in the quantitative numbers for this image (Depth Anything $\delta_1 = 99.95\%$, 
  Marigold $\delta_1 = 99.88\%$ and Ours $\delta_1 = 99.18\%$). The reason is the sparse ground truth LIDAR data with holes for objects with fine details (see bottom, left). 
  While the data-driven method, Depth Anything, requires a large corpus of training data (1.5M labelled and 62M unlabelled images), ours 
  and Marigold only need 74K synthetic training images.
  \vspace{-0.5cm}}
  \label{fig:intro_teaser}
\end{figure}

\section{Introduction} 
\label{sec:intro}

Depth estimation from a single image is a long-standing problem in computer vision. Although it is formally ill-posed, metric and affine-invariant depth estimation are well-defined tasks, and humans can also solve it approximately from cues such as texture-gradient, shading and shadows, or relative size~\cite{palmer1999visionscience}. 
Monocular-image depth estimation can also serve as an essential component for monocular-video depth estimation of dynamic scenes~\cite{cheng2024afnet} or in case of a forward-moving camera, as in a famous ``Hitchcock shot''. 

In the area prior to deep learning, depth estimation was only performed on a coarse superpixel-resolution, e.g. \cite{Saxena_maked3D}, or by approximating the scene with vertical planes, as in the automatic photo pop-up work~\cite{Hoiem2005_photoPopUp}. \footnotetext{With precomputed captions. On the fly, BLIP-2~\cite{BLIP2_Krause_2023_PLMR} image captioning adds on average additional 0.13 sec in runtime.} In recent years, the research field underwent a major transformation and current methods achieve a level of detail that is oftentimes not even captured in the respective ground truth data, due to physical limitations of depth measurement devices, such as LIDAR. 


In this work, we address the task of zero-shot, monocular depth estimation. This means that, once trained, our approach should work well on a broad range of image domains, such as indoor and outdoor, as well as in a variety of conditions, such as imagery of night scenes. Arguably, the first work that attempted zero-shot depth estimation has been MiDaS~\cite{ranftl2020midas}. The first MiDaS version uses a high-capacity encoder, which got replaced by a Vision-Transformer~\cite{ViT_Dosovitskiy_2021_ICLR} later on (MiDaS v3.0~\cite{Ranftl2021DPT}). The key strategy to success has been to use a very large corpus of labelled training images, real and synthetic, from different domains. The latest work with this data-driven philosophy is Depth Anything~\cite{Yang2024_DepthAnything}, which achieves impressive results using $1.5$M labelled and additionally $62$M unlabelled training images. It is a transformer-based architecture which is very fast at test time. \cref{fig:intro_teaser} shows a result of Depth Anything~\cite{Yang2024_DepthAnything} for a $768 \times 512$ image. %

Given the rise of generative Text-to-Image Diffusion models, with Stable Diffusion (SD)~\cite{LDM_Rombach_2022_CVPR} as one of the first foundation models, a new paradigm for monocular depth estimation has emerged. It has been observed in various works~\cite{liu2024understanding, tang2023correspondence} that SD ``understands the intrinsic representation of the real 3D world'' to some extent in order to generate images. For instance, in cross- and self-attention maps single objects are visible.
Equipped with this insight, various works~\cite{Marigold_Ke_2024_CVPR, VPD_Zhao_2023_ICCV, DMP_Lee_2024_CVPR, TADP_Kondapaneni_2024_CVPR, ECoDepth_Patni_2024_CVPR, MetaPrompt_Wan_2023_arXiv} have used the power of SD, or other diffusion models, to perform depth estimation from a given image. There are, in general, two lines-of-work for doing so. The most prominent line-of-work~\cite{Marigold_Ke_2024_CVPR, DMP_Lee_2024_CVPR} takes a diffusion model 
and ``repurposes'' it for the task of depth estimation, such as denoising a random noise-image to a depth-image output, guided by the given image. 
In contrast to pure data-driven approaches, it estimates a more detailed depth and, according to \cite{DMP_Lee_2024_CVPR}, also generalises better to very different domains such as art paintings. Furthermore, since the knowledge of ``realism'' originates from the diffusion model, it can be trained well with as little as 74K synthetic training images, e.g. \cite{Marigold_Ke_2024_CVPR}, which is only $4.9\%$ of labelled training images compared to the data-driven Depth Anything~\cite{Yang2024_DepthAnything} method. \cref{fig:intro_teaser} shows a result from Marigold~\cite{Marigold_Ke_2024_CVPR}, which is, arguably, the leading method of this line-of-work.

However, this line-of-work has two major drawbacks: (i) It is rather slow, since it requires multiple denoising steps to map random noise to a depth map. Furthermore, due to the probabilistic nature, Marigold~\cite{Marigold_Ke_2024_CVPR} has to average over an ensemble output; (ii) If a latent Diffusion Model is used, e.g. SD, the depth-based loss function has to be defined in latent domain. We show later that this leads to sub-optimal performance. These two drawbacks can be redeemed by simply using SD solely as a feature representation, i.e the features of a {\em single} denoising step of SD are extracted and fed into a depth prediction module. This marks the second line-of-work~\cite{VPD_Zhao_2023_ICCV, TADP_Kondapaneni_2024_CVPR, ECoDepth_Patni_2024_CVPR, EVP_Lavreniuk_2023_arXiv} which, to the best of our knowledge, has so far not been applied to zero-shot depth estimation. Indeed, when integrating SD-features in a straightforward manner to a downstream task then the performance is considerably below state-of-the-art, as shown later. The focus of this work is to look into the details of {\em what} should be extracted from SD and {\em how} it should be integrated into a downstream network. This analysis pays-off, and we arrive at a method that is quantitatively marginally superior to the leading diffusion-based method, Marigold~\cite{Marigold_Ke_2024_CVPR}, while being on average over $100\times$ faster. When comparing with Marigold, our results are also more robust to extreme imaging conditions such as night scenes with very limited illumination. We conjecture that this is due to Marigold changing the weights of the Stable Diffusion prior while we keep the rich representation unchanged. We call our method PrimeDepth, and show a result 
in \cref{fig:intro_teaser}. To summarise, our contributions are as follows: 
\begin{itemize}
    \item Preimage, an image representation of feature maps and attention maps that are derived from the single, last denoising step of Stable Diffusion~\cite{LDM_Rombach_2022_CVPR}. The preimage can effectively be used for a downstream task when combined with a refiner network that has an inductive bias by design.
    \item PrimeDepth, a single-step diffusion-based approach for zero-shot depth estimation. It has two conceptual advantages over multi-step diffusion approaches: i) very fast inference, e.g. on average $100\times$ faster than Marigold~\cite{Marigold_Ke_2024_CVPR}; ii) utilizing a loss function in pixel domain and not latent domain.     
    \item PrimeDepth is in our quantitative evaluation the runner-up, and only inferior to Depth Anything~\cite{Yang2024_DepthAnything} that is, however, a data-driven approach and requires over $20 \times$ more labelled training data, as well as roughly $25 \times$ longer training times based on the information provided in \cite{Yang2024_DepthAnything}. PrimeDepth also predicts more detailed depth maps than Depth Anything and is more robust than Marigold~\cite{Marigold_Ke_2024_CVPR} for challenging scenes such as at nighttime. 
\end{itemize}

\section{Related Work} 
\label{sec:related_work}

\subsection{Denoising Diffusion Models}

Since their introduction~\cite{DM_Origin, DDPM, song2021scorebased}, diffusion models have vastly advanced the field of unconditional and conditional image generation~\cite{Ho2022, hu2023self, NEURIPS2021_49ad23d1}. 
By construction, these models have the theoretical capability to represent any arbitrary data distribution~\cite{DM_Origin}. 
This makes them the spearhead in Text-to-Image (T2I) generation~\cite{LDM_Rombach_2022_CVPR, Dalle3, zheng2024cogview3, Podell2023_SDXL} and tasks requiring a rich and faithful image representation, such as image editing~\cite{hu2024instruct, goel2023pair, couairon2022diffedit, choi2023custom} or image-to-image translation~\cite{saharia2022palette, wang2022pretraining}. 
Many works have been dedicated to speed-up the generation process starting from optimised sampling~\cite{DDIM, dockhorn2022genie, lu2022dpm, zhang2022fast} up to the distillation of the internal representation of pre-trained diffusion models into single step methods~\cite{sauer2023adversarial, lin2024sdxl_lightning, yin2024one}.
In this work, we use the rich representation of the pre-trained Stable Diffusion and apply it to the task of zero-shot monocular depth estimation using only one single iteration.

\subsection{Zero-Shot Monocular Depth Estimation}

 In contrast to earlier works, which addressed in-domain monocular depth estimation~\cite{eigen2014KITTI_Eigensplits, fu2018DORN, bhat2021adabins, lee2019bts, garg2016unsupervisedgeometry,VPD_Zhao_2023_ICCV}, zero-shot depth estimators aim to generalise to unseen data distributions. Although there have been efforts on ordinal~\cite{chen2016DIW, xian2018redweb} and metric depth estimation~\cite{yin2023metric3d}, in this work, we focus on the affine-invariant setting~\cite{ranftl2020midas}. Thus far, two main research directions have been explored. The first leverages strong priors, such as T2I diffusion models (see \cref{subsec:diffusion_based_mde}), while the second, discussed in this section, is data-driven~\cite{ranftl2020midas, Ranftl2021DPT, eftekhar2021omnidata, yin2021LeRes, zhang2022HDN, Yang2024_DepthAnything}. MiDaS~\cite{ranftl2020midas} proposes training on diverse datasets, while DPT~\cite{Ranftl2021DPT} and Omnidata~\cite{eftekhar2021omnidata} further expand the number of training data and use transformers. LeReS~\cite{yin2021LeRes} proposes a framework that can additionally estimate the depth-shift and focal length, while HDN~\cite{zhang2022HDN} improves on detailed depth prediction with a multi-scale depth normalisation method. Finally, Depth Anything~\cite{Yang2024_DepthAnything} expands the datasets to large-scale unlabelled data with a challenging student-teacher optimisation target. Since Depth Anything is, to the best of our knowledge, quantitatively the leading method in this research direction, we perform a detailed comparison to it.

\subsection{Diffusion-Based Monocular Depth Estimators}
\label{subsec:diffusion_based_mde}
Diffusion models for depth prediction can be roughly assigned to three different paradigms.

\textbf{Diffusion paradigm} includes methods that take a not pre-trained diffusion framework and train it from scratch to denoise depth maps~\cite{DDP_Ji_2023_ICCV, DDVM_Saxena_2023_NIPS, DepthGen_Saxena_2023_arXiv, DMD_Saxena_2023_arXiv, DiffusionDepth_Duan_2023_arXiv}.
DepthGen~\cite{DepthGen_Saxena_2023_arXiv} conditions diffusion on an RGB image and operates in pixel domain. The authors addressed training on incomplete noisy ground truth data using simple interpolation and 
unsupervised auxiliary pre-training. DDVM~\cite{DDVM_Saxena_2023_NIPS} extended pre-training to include synthetic data and reduced the computational overhead through coarse-to-fine refinement in a patch-wise manner. DMD~\cite{DMD_Saxena_2023_arXiv} extends DDVM to a joint indoor-outdoor predictor by learning depth in log space.
DDP~\cite{DDP_Ji_2023_ICCV} and DiffusionDepth~\cite{DiffusionDepth_Duan_2023_arXiv} encode the conditional image with an off-the-shelf feature extractor to a smaller resolution prior to diffusion. 
Furthermore, such methods yield a posterior uncertainty by design based on unstable pixel regions~\cite{DepthGen_Saxena_2023_arXiv, DDP_Ji_2023_ICCV, DDVM_Saxena_2023_NIPS, DMD_Saxena_2023_arXiv}.

\textbf{Pre-trained diffusion paradigm} uses a pre-trained network, oftentimes a T2I model, and fine-tunes it on the downstream task~\cite{Marigold_Ke_2024_CVPR, DMP_Lee_2024_CVPR}. This paradigm has recently garnered popularity with the release of Marigold~\cite{Marigold_Ke_2024_CVPR}. Marigold performs diffusion and depth predictions in latent domain and is trained exclusively on limited synthetic data. Its main drawback is the computational overhead with a multi-step denoising approach, combined with an ensemble scheme.
Due to conceptual similarity, we compare extensively to Marigold.   
The goal of DMP ~\cite{DMP_Lee_2024_CVPR} is to show the generalisation of Stable Diffusion~\cite{LDM_Rombach_2022_CVPR} for downstream tasks such as depth estimation of highly creative synthetic data. It is done by training a LoRA~\cite{LoRA_Hu_2022_ICLR} and reformulating denoising as blending the image and depth map to achieve deterministic prediction. This paradigm demonstrates good zero-shot performance with detailed predictions, however, it is slow at inference time.

\textbf{Representation extractor paradigm}, in contrast to the previous, considers the pre-trained T2I network primarily as a feature representation akin to traditional backbones~\cite{VPD_Zhao_2023_ICCV, EVP_Lavreniuk_2023_arXiv, TADP_Kondapaneni_2024_CVPR, DatasetDM_Wu_NIPS_2023, ECoDepth_Patni_2024_CVPR, MetaPrompt_Wan_2023_arXiv}. The representation is extracted using image inversion and fed to a trainable depth predictor. Our method, PrimeDepth, follows this paradigm. 
For depth estimation, the pioneering work VPD~\cite{VPD_Zhao_2023_ICCV} extracts intermediate image features at four resolutions from Stable Diffusion~\cite{LDM_Rombach_2022_CVPR} and fuses them into a single low-resolution feature block using a semantic Feature Pyramid Network (FPN)~\cite{FPN_Kirillov_2019_CVPR}. The block is then processed by a lightweight upsampling decoder to form the final dense prediction. Notably, VPD demonstrated that diffusion models provide a competitive representation that surpasses traditional visual backbones like SwinV2~\cite{SwinV2_Liu_2022_CVPR}.
VPD considered only the task of in-domain monocular depth estimation.
In order to still compare to VPD, we build a VPD-like architecture within our framework and show that it is inferior.   
Recently, TADP~\cite{TADP_Kondapaneni_2024_CVPR} investigated the text guidance employed by VPD, showing that text-image alignment promotes better performance on downstream tasks. In their in-depth analysis of various captioning approaches they found that automated image captioning such as BLIP-2~\cite{BLIP2_Krause_2023_PLMR} yield best performance. 
A concurrent work, ECoDepth~\cite{ECoDepth_Patni_2024_CVPR} claimed that text guidance lacks expressiveness for perception tasks and replaced the text embeddings with ViT~\cite{ViT_Dosovitskiy_2021_ICLR} embeddings extracted from the input image. Finally, MetaPrompts~\cite{MetaPrompt_Wan_2023_arXiv} completely obviates the need for captions by making the text embeddings trainable. In our work, we use BLIP-2 for caption generation, since improving text embeddings is not our focus. The big advantage of this paradigm compared to the previous is inference speed, as it requires only one single network forward pass.
%
\vspace{-0.2cm}
\section{Method} 
\label{sec:method}

\subsection{Stable Diffusion} 
\label{subsec:stable_diffusion}

As a latent Text-to-Image (T2I) diffusion model, Stable Diffusion (SD) generates images by successively transforming Gaussian noise $z_T$ to a latent encoding $z_0$ in 1000 overall steps. This generative process in latent domain is performed by a \mbox{U-Net} $f$ consisting of different neural blocks, \ie residual, self- and cross-attention blocks, that are executed sequentially and at different scales. The excessive training of SD on the diverse LAION-5B~\cite{Schuhmann2022_LaionAE} dataset allows for a rich representation of arbitrary imagery. The motivation for our SD representation stems from~\cite{luo2024diffusionTimestepFeatures, hedlin2024timestepFeatures2}, which has shown that for the semantic correspondence task, the last denoising step $f(z_1)$ contains rich information. This intuitively makes sense since at that time-step the representation contains the least amount of ``potential hallucinations'' of the generative model.

\subsection{Preimage Representation}
\label{subsec:feature_extraction}
\begin{figure}[tb]
  \centering
  \includegraphics[width=.98\linewidth]{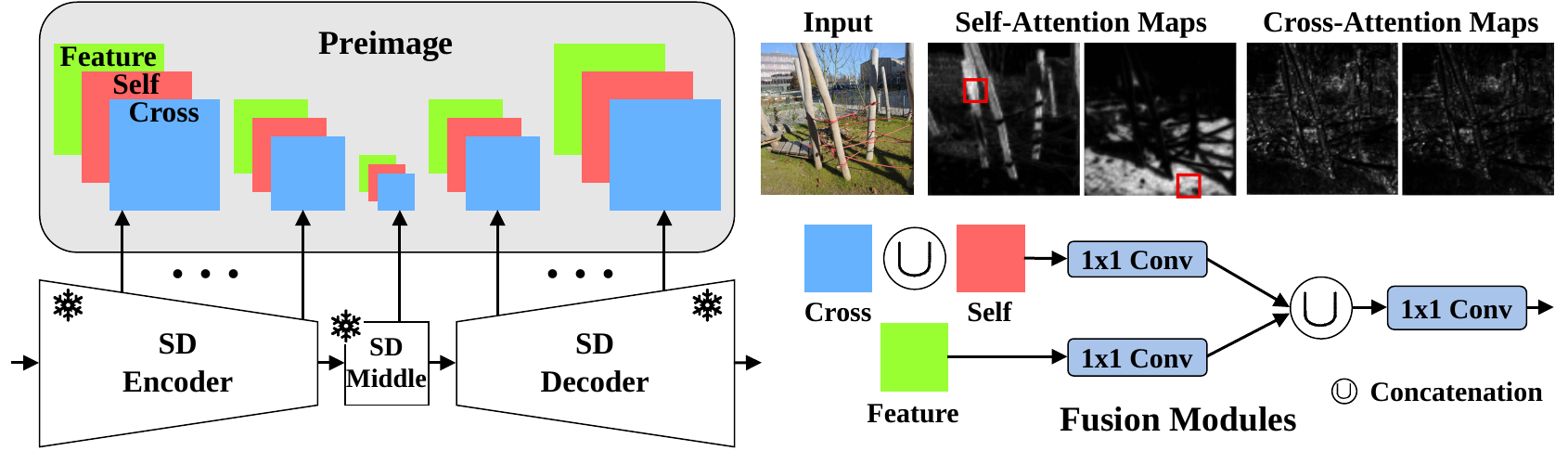}
  \caption{(Left) Stable Diffusion preimage consisting of intermediate feature maps, cross- and self-attention maps for every neural block of the last denoising step. (Right) Examples of self-attention maps, with respect to the red square, and cross-attention maps. Below is the fusion model shown, for the aggregation of the preimage parts.
  \vspace{-0.5cm}}
  \label{fig:pre_image_and_attentions}
\end{figure}

We define our preimage as all the intermediate, multi-scale feature maps, cross- and self-attention maps of every neural block in the last denoising step of SD version 2.1, see \cref{fig:pre_image_and_attentions} (left).
In contrast to prior work VPD~\cite{VPD_Zhao_2023_ICCV}, we also utilise
self- and cross-attention maps which encode similar but complementary learned structural and semantic concepts. While the feature activation maps can be used right away, the attention maps require further processing to reduce the respective channel-size and possible redundancies. With $\mathcal{H}$ attention heads, height $h$ and width $w$, the self-attention maps have the dimension $\mathcal{H} \times h \times w \times N_p$, with $N_p=h\cdot w$ being the number of pixels, since all pixels attend to each other. Similarly, the cross-attention maps, attending to 77 text tokens, have the dimension $\mathcal{H} \times h \times w \times 77$. For an image with the resolution of $512 \times 512$, the corresponding largest self-attention map in latent domain has the dimension $\mathcal{H} \times 64 \times 64 \times 4096$. Using all maps is not only computationally infeasible, but also impossible, as the channel size $N_p$ changes with the resolution of the input. We circumvent both problems by dividing the attention maps in $8 \times 8$ regions and averaging over the respective attention-regions. Because each head can potentially attend to, and therefore encode, different kind of content, we do not average over heads. This means formally,
\begin{align}
        \mathbb{R}^{\mathcal{H} \times h \times w \times N_p} \xrightarrow[\text{average}]{\text{regional}} \mathbb{R}^{\mathcal{H} \times h \times w \times 64} \longrightarrow \mathbb{R}^{h \times w \times (64 \cdot \mathcal{H})} \qquad & \text{(self-attention)} \\
        \mathbb{R}^{\mathcal{H} \times h \times w \times 77}  \longrightarrow \mathbb{R}^{h \times w \times (77 \cdot \mathcal{H})} \qquad & \text{(cross-attention).}
\end{align}
\cref{fig:pre_image_and_attentions} (right) shows examples for attention maps after averaging over regions.
Before passing the extracted parts of the preimage to the refining network, the features need to be aggregated. The fusion module for the aggregation is illustrated in \cref{fig:pre_image_and_attentions} (bottom, right) and comprises a sequence of concatenations and convolution projections of all feature and attention maps.

\subsection{PrimeDepth Architecture} 
\label{subsec:network_design}

\begin{figure}[tb]
  \centering
  \includegraphics[width=.9\linewidth]{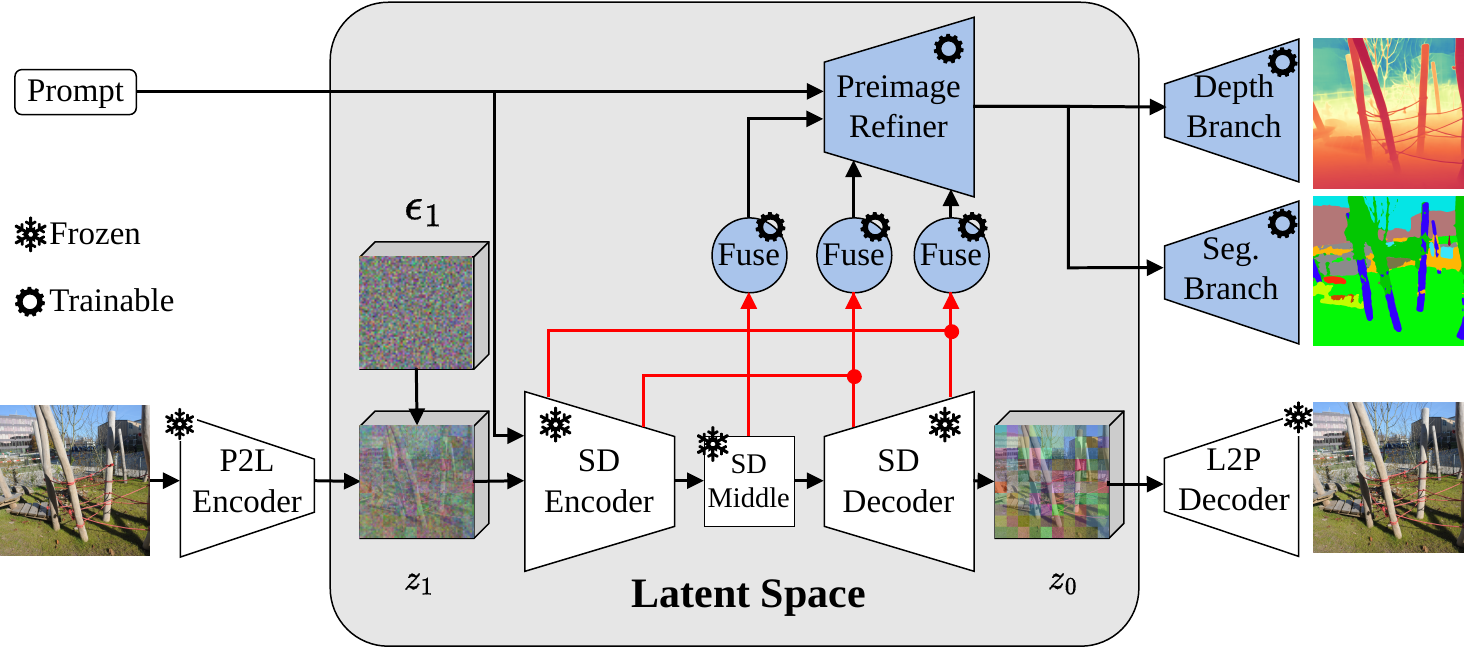}
  \caption{PrimeDepth Pipeline. The input image is first encoded to latent domain, augmented with one noise step and processed by the frozen U-Net of Stable Diffusion. The intermediate parts of the preimage (red arrows) are aggregated with the fusion module (see \cref{fig:pre_image_and_attentions}) and provided to the preimage refiner network at the respective intermediate stages. The output of the refiner is fed to two downstream heads for the respective downstream tasks. 
  \vspace{-0.5cm}}
  \label{fig:network_design}
\end{figure}

The full architecture design of PrimeDepth is shown in \cref{fig:network_design}. 
The depth prediction pipeline starts by encoding the image $x$ with the SD Pixel-to-Latent (P2L) encoder to its latent domain representation $z_0$ and adding one step of noise to receive the slightly noisy latent $z_1$. In the denoising step of the frozen U-Net, the preimage parts, shown with red arrows, are passed through fusion modules (\cref{fig:pre_image_and_attentions}, bottom right)  and provided to the preimage refining decoder via concatenations at the respective blocks.
The output of the refining decoder is a latent representation that can be used by downstream decoder-heads to get the final output in pixel domain. In  contrast to diffusion-based approaches that use multiple-denoising steps, like Marigold~\cite{Marigold_Ke_2024_CVPR}, we define the loss function in the output domain. The downside of defining the loss function in latent domain like Marigold is twofold: i) The task decoder cannot be trained at the same time; ii) The loss function in latent domain is a surrogate of the one in pixel domain. %

The downstream heads are trainable copies of the SD latent-to-pixel (L2P) decoder with an increased bottleneck size from 4 input channels to 512. We use two downstream heads for training, one for depth prediction and the other one for semantic segmentation as regularisation. We utilise 150 semantic classes as in \cite{zhou2017Ade20k} and see later that it improves the performance.
Note that previous works have also demonstrated the benefits of auxiliary tasks for regularisation, including semantics~\cite{chen2019towards, guizilini2020semantically, klingner2020self, xu2022mtformer} or normal prediction~\cite{long2021adaptive, yin2019enforcing}.

To summarise, PrimeDepth fulfils three conceptual requirements. \textit{(i) The base model weights remain unaltered.} We believe that this is an advantage, since SD was trained extensively on a large amount of data. By adapting the weights, as e.g. done in Marigold~\cite{Marigold_Ke_2024_CVPR} and VPD~\cite{VPD_Zhao_2023_ICCV}, the representation capabilities may decrease and with this the generalisation power, as seen later in reduced robustness of e.g. Marigold.  
\textit{(ii) Utilisation of the full preimage representation.} As we see later, using the full representation of SD, i.e. using all intermediate feature maps, self- and and cross-attention maps, is beneficial for the depth prediction task.  
\textit{(iii) Exploiting architectural inductive bias for the preimage representation.} The refining decoder has the same architecture as the SD decoder, but with only 50\% of the original channel sizes. Hence it introduces a natural inductive bias to the architecture for the gradual processing of the preimage. All components of the preimage are provided to the refiner at its respective resolution stages.

It is worth to note that using an inductive architectural bias is not new, and for instance has shown to be beneficial in the context of controlled image generation for SD~\cite{zavadski2023controlnet-xs}. We believe that transformer architectures could also benefit from the integration of the generative preimage with architectural bias and sketch a possible integration to a DPT-based~\cite{Ranftl2021DPT} refiner in the supplement.

\subsection{Training Protocol} 
\label{subsec:training_protocol}

Although the preimage refiner network has the same architecture as the SD-decoder, it has to be initialised with random weights since the channel sizes differ. 
To compensate for the random initialisation, we first pre-train the model with unlabelled data. Note that this is a common strategy, also used for instance by Depth Anything~\cite{Yang2024_DepthAnything}, and gives a marginal quantitative boost in performance as shown later. 
For the pre-training, we use a subset of 600K images from the LAION-Aesthetics dataset, which contains images of high visual quality from arbitrary domains. We compute a pseudo ground truth for depth using Depth Anything~\cite{Yang2024_DepthAnything} and for segmentation maps with \mbox{InternImage-H}~\cite{wang2022internimage}. We use \mbox{BLIP-2}~\cite{BLIP2_Krause_2023_PLMR} to captionise all images. The model is first pre-trained on the pseudo ground truth data and then trained on 74K images of synthetically labelled data. Since our model uses a preimage that is computed in one single forward pass, we are not restricted by latent losses. We use the scale and shift invariant loss of MiDaS~\cite{ranftl2020midas} for the depth estimation in pixel domain. We first shift and scale the ground truth depth
        $d^\ast = \frac{d - t(d)}{s(d)}$ with $t(d) = \text{median}(d)$ and $s(d) = \frac{1}{hw} \sum^{hw}_{i=1} |d_i - t(d)|$ and then compute the mean squared error between the affine-invariant ground truth $d^\ast$ and the depth prediction $\hat d$, which is aligned to $d^\ast$ by least squared errors $\alpha(d^\ast, \hat d)$, to get the shift and scale invariant loss
    \begin{align}
        \mathcal{L}_{\text{ssi}} = \text{MSE}\left(d^\ast, \alpha(d^\ast, \hat d)\right).
    \end{align}
For the regularisation through semantic segmentation, we use a combination of the Dice Loss~\cite{milletari2016DiceLoss} and the Focal Loss~\cite{lin2017FocalLoss} with respect to the one-hot encoded semantic segmentation maps for $C=150$ classes

    {\small 
    \begin{align}
        \mathcal{L}_{\text{dice}} = 1 - \frac{1}{C} \sum_{c=0}^{C-1} \frac{2 \sum_{i=1}^{hw} y_i^{\ast c}\cdot \hat y^c_i }{\sum_{i=1}^{hw} (y_i^{\ast c} + \hat y_i^c)},   \qquad
        \mathcal{L}_{\text{focal}} = - \sum_{i=1}^{hw} (1 - y^\ast_i \cdot \hat y_i)^2\log(y^\ast_i \cdot \hat y_i)
    \end{align}
    }

with $y^\ast, \hat y$ being the ground truth and the predicted segmentation maps, respectively.
We dynamically weigh the depth loss to the segmentation loss and arrive at the final training objective
    \begin{align}
        \mathcal{L} = \mathcal{L}_{\text{dice}} + \mathcal{L}_{\text{focal}} + \lambda \mathcal{L}_{\text{ssi}}, \quad \text{with}\quad \lambda = \text{sg}\left(\frac{\mathcal{L}_{\text{dice}} + \mathcal{L}_{\text{focal}}}{\mathcal{L}_{\text{ssi}}} \right),
    \end{align}
where sg($\cdot$) represents the stop-gradient operator.

\vspace{-0.3cm}
\section{Experiments} 
\label{sec:experiments}

\subsection{Datasets and Evaluation Metrics} 

\textbf{Training datasets.} We utilised the same datasets as Marigold~\cite{Marigold_Ke_2024_CVPR} to allow for a fair comparison. These are the two synthetic and densely-labelled datasets Hypersim~\cite{roberts_Hypersim} and Virtual KITTI 2~\cite{cabon2020vkitti2}, which represent indoor and outdoor scenarios. The LAION-Aesthetics dataset~\cite{Schuhmann2022_LaionAE} is used for pre-training our method, see details in \cref{subsec:training_protocol}, and gives a marginal boost in performance, see below. 

\textbf{Test datasets.} We evaluate in total on five unseen datasets: KITTI~\cite{geiger2012KITTI} using Eigen split~\cite{eigen2014KITTI_Eigensplits} (654 images), NYUv2~\cite{NYUv2_Silberman_2012_ECCV} (656 images), ETH3D~\cite{schops2017ETH3D} (456 images), rabbitai~\cite{schilling2020rabbitai} (60 images)\footnote{See Robust Vision Challenge 2020 (http://www.robustvision.net/rvc2020.php).}, and a manually chosen subset of nuScenes~\cite{caesar2020nuscenes}, we term nuScenes-C (3489 images). Since KITTI, ETH3D and NuScenes-C were captured with LIDAR sensors, the ground truth measurements are sparse with occasional holes, \eg in the sky and objects with fine details. The NYUv2 dataset was captured with an active RGBD-kinect sensor. It has dense ground truth apart from occasionally missing regions at depth-borders. While the rabbitai dataset has fewest number of images, it represents, at the same time, a dataset with dense and highly detailed ground truth depth. It was captured with a passive light-field camera setup. %
We created the nuScenes-C dataset by selecting 87 scenes (in total 3489 images) from nuScenes that represent challenging scenarios, such as images in bad weather conditions, nighttime, and adversarial visual cues including reflections and artefacts on the lens. We further categorised the nuScenes-C dataset into 6 categories, see \cref{fig:challenging_scenarios_boxplot}. 

\begin{figure}[tb!]
  \centering
  \includegraphics[width=.98\linewidth]{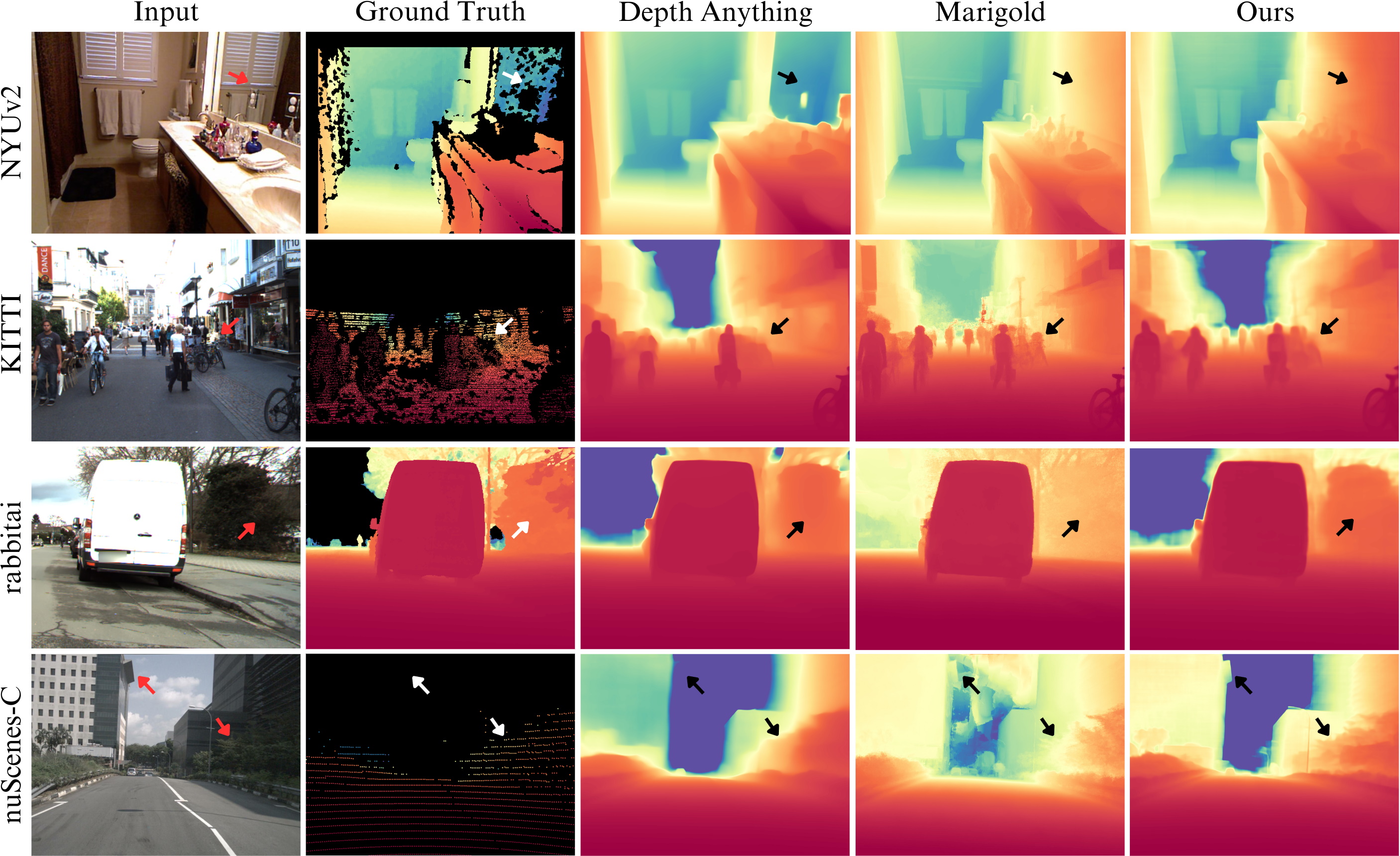}
  \caption{Qualitative results of two competing methods (Depth Anything~\cite{Yang2024_DepthAnything} and Marigold~\cite{Marigold_Ke_2024_CVPR}) for 4 datasets, while results for ETH3D are shown in \cref{fig:intro_teaser}. The main visual artefacts of the respective methods are indicated by arrows. The prominent observations across many images are as follows. Depth Anything has less sharp depth maps (KITTI, nuScenes-C, \cref{fig:intro_teaser}) and can see inside mirrors (NYUv2) and through transparent surfaces (supplement). Marigold predicts sharp depth maps but sometimes with grainy artefacts (KITTI, \cref{fig:intro_teaser}). 
  It also struggles to predict sky (KITTI, nuScenes-C) and objects at mid-distance (rabbitai). Our method gives sharper depth maps than Depth Anything (KITTI, NuScenes-C), but can also struggle with sky (supplement).    
  \vspace{-0.6cm}
}
  \label{fig:main_qualitative}
\end{figure}

\textbf{Metrics.} As in training, a predicted depth map $\hat d$ is first affinely aligned with ground truth $d^*$. We use two common metrics: i) absolute relative error $AbsRel = \frac{1}{hw}\sum_{hw} \frac{|d^{*} - \hat{d}|}{d^{*}}$ and ii) accuracy under threshold $\delta_{1} = \frac{1}{hw}\sum_{hw}[max(\frac{\hat{d}}{d^{*}}, \frac{d^{*}}{\hat{d}}) < 1.25]$, measuring the fraction of pixels that satisfy the condition. 

\begin{table}[tb]
  \caption{Comparison of 8 methods for \textit{zero-shot} monocular depth estimation, where numbers are taken from Marigold~\cite{Marigold_Ke_2024_CVPR}, apart from the numbers of Depth Anything~\cite{Yang2024_DepthAnything}, taken from their article, and our numbers using our protocol. 
  Since the methods may not be evaluated consistently, the average rank may be the best performance indicator.}
  \label{tab:quantitative_external}
  \centering
  \resizebox{\textwidth}{!}{
  \begin{tabular}{lccccccccccc} 
    \toprule
    & \multicolumn{2}{c}{KITTI}
    & \multicolumn{2}{c}{NYUv2} 
    & \multicolumn{2}{c}{ETH3D} 
    & Average
    & Num.
    & Architecture
    \\
    Model     
    &    $\delta_1 \uparrow$    &    AbsRel $\downarrow$       
    &    $\delta_1 \uparrow$    &    AbsRel $\downarrow$    
    &    $\delta_1 \uparrow$    &    AbsRel $\downarrow$     
    & Rank
    & Train Samples
    & Type
    \\
    \midrule
    MiDaS~\cite{ranftl2020midas}           
                                & 63.0    
                                & 23.6    
                                & 88.5    
                                & 11.1    
                                & 75.2    
                                & 18.4    
                                & 8.0    
                                & 2M       
                                & ResNeXt101
                                \\
    LeReS~\cite{yin2021LeRes}               
                                & 78.4    
                                & 14.9    
                                & 91.6    
                                & 9.0    
                                & 77.7    
                                & 17.1    
                                & 6.5    
                                & 354K       
                                & ResNeXt101
                                \\
    Omnidata~\cite{eftekhar2021omnidata}    
                                & 83.5    
                                & 14.9    
                                & 94.5    
                                & 7.4    
                                & 77.8    
                                & 16.6    
                                & 5.7    
                                & 12.2M       
                                & Transformer
                                \\
    DPT~\cite{Ranftl2021DPT}                      
                                & 90.1    
                                & 10.0    
                                & 90.3    
                                & 9.8    
                                & 94.6    
                                & 7.8    
                                & 4.7    
                                & 1.4M       
                                & Transformer
                                \\
    HDN~\cite{zhang2022HDN}                    
                                & 86.7    
                                & 11.5    
                                & 94.8    
                                & 6.9    
                                & 83.3    
                                & 12.1    
                                & 4.5    
                                & 300K       
                                & Transformer
                                \\
    Marigold~\cite{Marigold_Ke_2024_CVPR}            
                                & 91.6    
                                & 9.9    
                                & 96.4    
                                & \underline{5.5}    
                                & \underline{96.0}    
                                & \textbf{6.5}    
                                & 2.3    
                                & 74K       
                                & Multi-step diffusion
                                \\
    Depth Anything~\cite{Yang2024_DepthAnything}       
                                & \textbf{94.7}    
                                & \textbf{7.6}    
                                & \textbf{98.1}    
                                & \textbf{4.3}    
                                & 88.2    
                                & 12.7    
                                & \underline{2.2}    
                                & 63.5M       
                                & Transformer
                                \\
    \midrule
    Our                 
                                & \underline{93.7}    
                                & \underline{7.9}    
                                & \underline{96.6}    
                                & 5.8    
                                & \textbf{96.7}    
                                & \underline{6.8}    
                                & \textbf{2.0}    
                                & 74K       
                                & Single-step diffusion
                                \\
  \bottomrule
  \end{tabular}
  \vspace{-0.2cm}
}
\end{table}

\vspace{-0.1cm}
\subsection{Comparison to other Methods} 

While reproducing the results of previous methods, it became apparent that the evaluation for the same datasets can unfortunately differ substantially between articles. As an example, Depth Anything~\cite{Yang2024_DepthAnything} reports $\delta_1 = 88.2\%$ for the ETH3D dataset, while using our protocol, which tries to match the numbers of Marigold~\cite{Marigold_Ke_2024_CVPR}, the score is much better, i.e. $\delta_1 = 98.1\%$ for ETH3D. \cref{tab:quantitative_external} shows a comparison of 8 methods for three unseen datasets, where all numbers are from Marigold, apart from the numbers of Depth Anything~\cite{Yang2024_DepthAnything} which are from their article and our numbers using our protocol. Since individual numbers might not be fully representative, we provide the average rank as an indicator for overall performance\footnote{We take the average with respect to both measures, $\delta_1$ and AbsRel.}. We observe that the two recent methods, Marigold and Depth Anything, have clearly the best rank, i.e. 2.3 and 2.2, alongside our approach with rank 2.0. Given this, we do a detailed analysis of these three methods for 5 unseen datasets using our evaluation protocol, see \cref{tab:quantitative_our_protocol}. We see that Depth Anything is superior to the other methods, while our method is clearly the runner-up. Our method without pre-training is on a par with Marigold. As mentioned in the introduction, Marigold and our approach only use a fraction of labelled training data compared to Depth Anything (74K versus 1.5M).
Furthermore, since our approach uses only a single diffusion step, we are on average 100 times faster than Marigold at test time.\footnote{Computed from 1000 runs of different image resolutions on an A100 GPU.} Despite the pre-training on Depth Anything predictions, the results of PrimeDepth and Depth Anything are complementary. To show this, we provide two simple approaches for combining the results of Depth Anything and PrimeDepth. The Pixel-wise-Average method simply averages the two results pixel-wise. It improves over both Depth Anything and PrimeDepth, meaning that the two signals are not fully correlated. We also run an Image-Oracle, which selects either the Depth Anything or PrimeDepth result. The oracle improves the results noticeably, and in $21\%$ to $55\%$ of cases the PrimeDepth result is selected.
Qualitative results are shown and discussed in \cref{fig:intro_teaser} and \cref{fig:main_qualitative}. In brief, the depth maps of Depth Anything exhibit blurriness, although this is not reflected in the quantitative results due to missing values in the ground truth. Also, Depth Anything has the same artefacts as in the ground truth, e.g. it sees through reflective objects. Marigold and our method are visually more detailed, however, again, giving missing ground truth it is difficult to clearly measure its correctness. Marigold is occasionally too detailed which results in grainy artefacts, and sometimes struggles to predict depth at mid-distance. Our method and Marigold sometimes do not predict sky at infinity.

\begin{table}[tb]
  \caption{Comparison to Depth Anything (DA) \cite{Yang2024_DepthAnything} and Marigold \cite{Marigold_Ke_2024_CVPR} for \textit{zero-shot} monocular depth estimation using the same (our) evaluation protocol.}
  \label{tab:quantitative_our_protocol}
  \centering
  \resizebox{\textwidth}{!}{
  \begin{tabular}{lccccccccccc} 
    \toprule
    & \multicolumn{2}{c}{KITTI}
    & \multicolumn{2}{c}{NYUv2} 
    & \multicolumn{2}{c}{ETH3D} 
    & \multicolumn{2}{c}{rabbitai} 
    & \multicolumn{2}{c}{nuScenes-C} 
    & Avg. Rank
    \\
    Model       
    &    $\delta_1 \uparrow$    &    AbsRel $\downarrow$       
    &    $\delta_1 \uparrow$    &    AbsRel $\downarrow$    
    &    $\delta_1 \uparrow$    &    AbsRel $\downarrow$    
    &    $\delta_1 \uparrow$    &    AbsRel $\downarrow$    
    &    $\delta_1 \uparrow$    &    AbsRel $\downarrow$    
    \\
    \midrule
    Image-Oracle (Upper Bound)                 
                                & {95.7} 
                                & {7.0} 
                                & {98.5} 
                                & 4.1 
                                & {98.4} 
                                & 4.9 
                                & {80.0} 
                                & {18.0} 
                                & {82.9} 
                                & {13.6} 
                                & -    
                                \\
    (Our\%/DA\%)            
                                & {(42/58)} 
                                & {(53/47)} 
                                & {(38/62)} 
                                & {(21/79)} 
                                & {(37/63)} 
                                & {(31/69)} 
                                & {(50/50)} 
                                & {(55/45)} 
                                & {(28/72)} 
                                & {(30/70)} 
                                & -    
                                \\
                                \midrule
    Depth Anything (DA)~\cite{Yang2024_DepthAnything}         
                                & \underline{94.6}    
                                & {8.0}    
                                & \textbf{98.0}    
                                & \textbf{4.3}    
                                & \textbf{98.1}    
                                & \underline{5.6}    
                                & \underline{76.9}    
                                & 20.7    
                                & \textbf{81.9}    
                                & \underline{14.5}    
                                & \underline{1.9}    
                                \\
    Marigold~\cite{Marigold_Ke_2024_CVPR}              
                                & 91.6    
                                & 10.0    
                                & 96.4    
                                & {5.5}    
                                & 96.5    
                                & {6.0}    
                                & 56.6    
                                & 27.2    
                                & 64.0    
                                & 24.1    
                                & 4.3    
                                \\
    Our w/o pre-train                      
                                & 91.2    
                                & 9.5    
                                & 91.8    
                                & 9.0    
                                & 95.1    
                                & 8.2    
                                & 71.6    
                                & {20.3}    
                                & 73.9    
                                & 18.5    
                                & 4.4    
                                \\
    Our                 
                                & {93.7}    
                                & \underline{7.9}    
                                & {96.6}    
                                & 5.8    
                                & {96.7}    
                                & 6.8    
                                & {76.2}    
                                & \underline{20.1}    
                                & {79.2}    
                                & {15.8}    
                                & 3.0    
                                \\
    Pixel-wise-Average (Our \& DA)                
                                & \textbf{95.3} 
                                & \textbf{7.3} 
                                & \underline{97.7} 
                                & \underline{4.6}  
                                & \textbf{98.1} 
                                & \textbf{5.5} 
                                & \textbf{77.7} 
                                & \textbf{19.4} 
                                & \underline{81.6} 
                                & \textbf{14.3} 
                                & \textbf{1.3}    
                                \\
  \bottomrule
  \end{tabular}
  }
  \vspace{-0.2cm}
\end{table}

\subsection{Comparison for challenging scenarios} 

\begin{figure}[t]
  \centering
  \includegraphics[width=.98\linewidth]{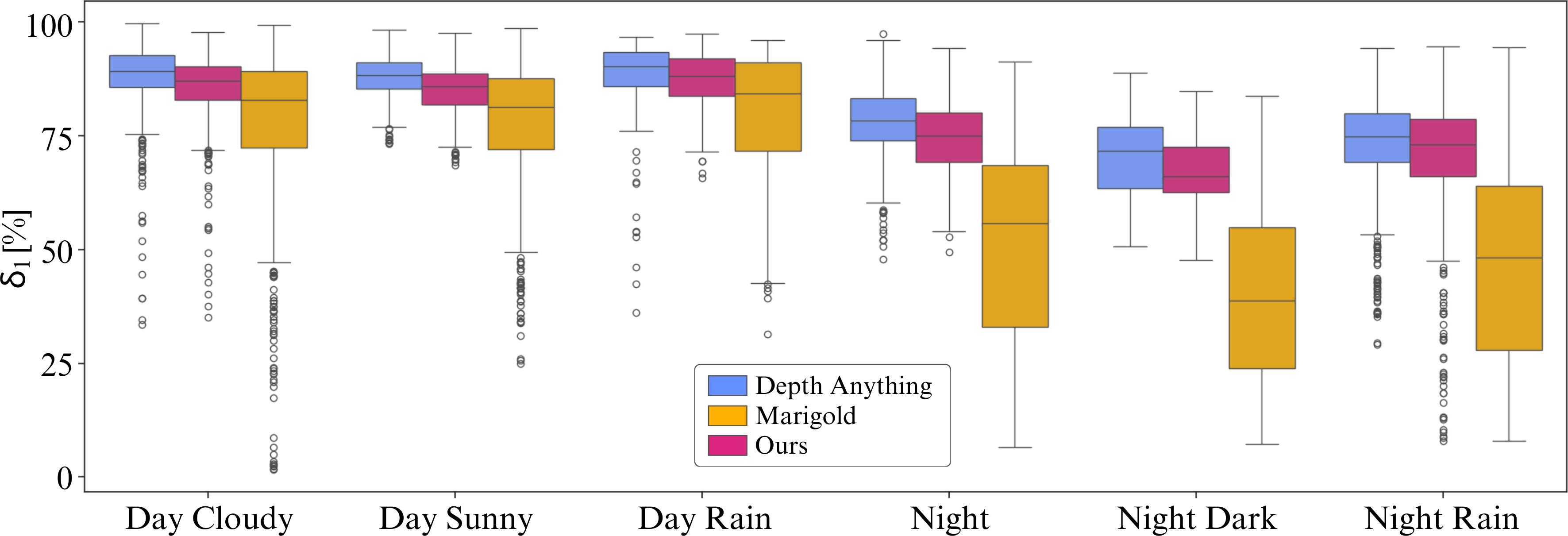}
  \caption{
  Box plot for the $\delta_1$ accuracy of challenging scenes from the nuScenes-C dataset, split into 6 categories, where long, vertical stripes provide the median values. Our PrimeDepth is consistently, marginally inferior to Depth Anything~\cite{Yang2024_DepthAnything}, but consistently and sometimes considerably superior to Marigold~\cite{Marigold_Ke_2024_CVPR}. The variability, measured by IQR score i.e. size of a box, is considerably higher for Marigold than Depth Anything and our method. For nighttime scenes, the performance drops for all methods, however Marigold is clearly more affected (lowest median and highest variability).
  \vspace{-0.5cm}
}
  \label{fig:challenging_scenarios_boxplot}
\end{figure}

\begin{figure}[tb]
  \centering
  \includegraphics[width=.98\linewidth]{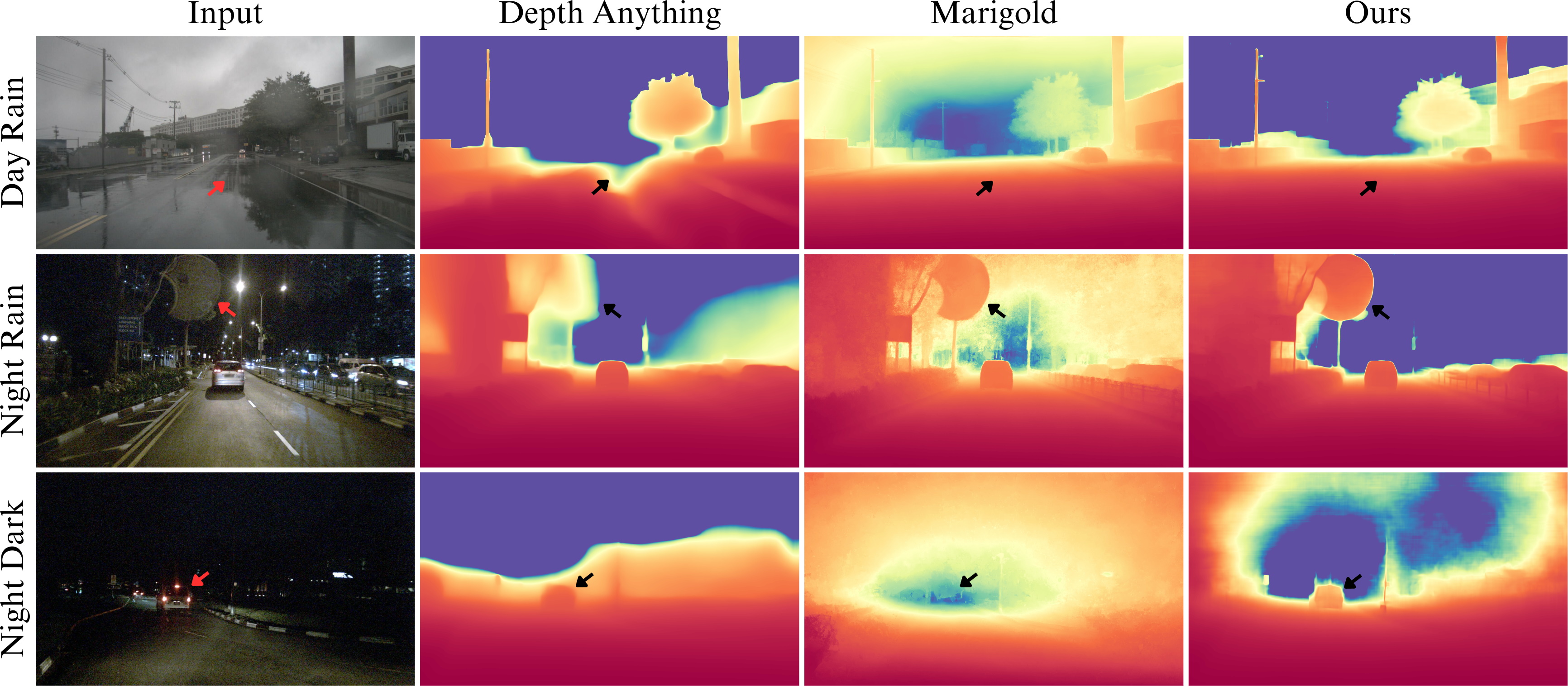}
  \caption{Challenging scenes for category ``Day Rain'' (top row), ``Night Rain'' (middle row) and ``Night Dark'' (bottom row). Depth Anything~\cite{Yang2024_DepthAnything} wrongly predicts depth for reflections on a wet surface (top row).  Marigold~\cite{Marigold_Ke_2024_CVPR} struggles to predict the depth of objects at mid-depth (bottom row) and sky (all rows). A failure case for both methods, Marigold and ours, is a water drop on the camera lens which is likely predicted as a round object in the scene (middle row).
  \vspace{-0.5cm}}
  \label{fig:challenging_scenarios_qualitative}
\end{figure}

To better understand the behavior of methods for challenging scenarios, we visualise their performance with respect to 6 different categories in form of a boxplot. \cref{fig:challenging_scenarios_boxplot} shows the $\delta_1$ accuracy computed on a per-frame basis. The median value is marked with a long, vertical stripe. The variability in the results is shown by the size of the solid box, which is the interquartile range (IQR) containing 50\% of all inlier predictions. 
We observe that the general trend matches that of the benchmark above. Our method is marginally inferior to Depth Anything~\cite{Yang2024_DepthAnything}, but consistently superior to Marigold~\cite{Marigold_Ke_2024_CVPR}. The IQR scores of our method and Depth Anything are similar and relatively small, in contrast to Marigold. For nighttime scenes, the drop in median score and higher IQR scores is noticeable for all methods, however, considerably and consistently more prominent for Marigold. We conjecture that the low robustness of Marigold stems from the fact that Stable Diffusion itself is modified from being an image generation method to being a depth prediction method. In contrast, we retain the rich information of the frozen Stable Diffusion model. Qualitative samples of challenging scenes at night and in rain are shown in \cref{fig:challenging_scenarios_qualitative}. It is noticeable that Marigold struggles to predict objects at mid-distance as well as the sky region. Depth Anything wrongly predicts depth for reflections on a wet surface. A failure case of both, Marigold and our approach, is a water drop on the camera lens, which is likely predicted as a round object in the scene. 

\vspace{-0.4cm}
\subsection{Ablation Study}
\label{subsec:ablation}
\vspace{-0.2cm}
To evaluate the effectiveness of the individual components of our method, we ablate in \cref{tab:abl_study} each design choice after pre-training on pseudo ground truth images only. The PrimeDepth architecture is called V.Full and performs best across the board. By training the model in latent domain (V.Latent), as done by methods like Marigold~\cite{Marigold_Ke_2024_CVPR}, we see a decrease in performance across all datasets. %
Omitting the regularisation with respect to the segmentation loss (V.NoSegmen), results in a similar decrease in the overall accuracy. Using V.NoSegmen and additionally removing either the feature maps (V.NoFeature) or the attention maps (V.NoAttent), gives a further decline in performance. Overall, feature maps appear to be more informative than attention maps, but attention maps still contain useful additional information. We also compare our network design with designs suggested in previous works~\cite{VPD_Zhao_2023_ICCV, TADP_Kondapaneni_2024_CVPR}. For this we use our full preimage and re-scale it as a single ``block'' with uniform resolution and provide it as regular input to the downstream task. The key difference to our design is that the natural inductive bias of the preimage refiner is not exploited. To be in spirit close to \cite{VPD_Zhao_2023_ICCV}, we inserted the preimage of V.NoSegmen as a block into our preimage refiner. We refer to this design as V.BlockPreIm (see full network design in supplement). The performance of V.BlockPreIm, compared to V.NoSegmen, is clearly inferior for the KITTI dataset and the densely labelled rabbitai dataset, while for the sparse nuScenes-C data, the performance is basically on par.

\begin{table}[tb]
  \caption{Ablation Study of different design choices, noted by Version.* (V.*). All versions are only pre-trained on a 600K image subset of LAION, and not trained on ground truth data as our variants in \cref{tab:quantitative_external} and \cref{tab:quantitative_our_protocol}.}
  \label{tab:abl_study}
  \centering
  \resizebox{\textwidth}{!}{
  \begin{tabular}{lccccccccccc} 
    \toprule
    &
    &
    &
    &
    & \multicolumn{2}{c}{KITTI} 
    & \multicolumn{2}{c}{rabbitai} 
    & \multicolumn{2}{c}{nuScenes-C} 
    & Avg. Rank
    \\
    Model
    & Segm
    & Atten
    & Feat
    & PixLoss
    &    $\delta_1 \uparrow$    &   AbsRel $\downarrow$     
    &    $\delta_1 \uparrow$    &   AbsRel $\downarrow$     
    &    $\delta_1 \uparrow$    &   AbsRel $\downarrow$     
    \\
    \midrule
    V.Full                 
                                &   x       
                                &   x       
                                &   x       
                                &   x       
                                & \textbf{92.8}     
                                &  \textbf{8.7}     
                                & \textbf{73.5}     
                                & \textbf{21.9}     
                                & \textbf{80.6}     
                                & \textbf{14.2}     
                                &    1.0 
                                \\
    V.Latent                 
                                &   x       
                                &   x       
                                &   x       
                                &   -       
                                & 92.5     
                                &  \underline{8.8}     
                                & 70.1     
                                & 24.4     
                                & 79.8     
                                & 15.0     
                                &  3.0 
                                \\
    V.NoSegmen                   
                                &   -       
                                &   x       
                                &   x       
                                &   x       
                                & \textbf{92.8}    
                                & 8.9    
                                & \underline{72.9}    
                                & 24.6             
                                & 79.8    
                                & 15.0    
                                &  3.0 
                                \\             
    V.NoAttent                 
                                &   -       
                                &   -       
                                &   x       
                                &   x       
                                & \underline{92.7}     
                                &  9.3      
                                & 72.7     
                                & 25.4     
                                & 79.1     
                                & 16.3     
                                & 4.2 
                                \\             
    V.NoFeature             
                                &   -       
                                &   x       
                                &   -       
                                &   x       
                                & 88.0     
                                & 11.4     
                                & 67.9     
                                & \underline{23.9}     
                                & 76.5    
                                & 16.9     
                                &  4.8 
                                \\
    V.BlockPreIm 
                                &   -       
                                &   x       
                                &   x       
                                &   x       
                                & 86.6     
                                & 12.4     
                                & 65.3     
                                & 26.5     
                                & \underline{80.1}     
                                & \underline{14.8}     
                                &  4.7 
                                \\
  \bottomrule
  \end{tabular}
  }
  \vspace{-0.0cm}
\end{table}

\vspace{-0.3cm}
\section{Conclusion and Future Work} 
\label{sec:conclusion}

In this work, we presented PrimeDepth, a method for harnessing and utilising the complete, rich preimage of Stable Diffusion for zero-shot monocular depth estimation. Our model achieves competitive results on a variety of datasets while being considerably faster than competitive diffusion-based approaches, like Marigold~\cite{Marigold_Ke_2024_CVPR}, owing to a single diffusion step. By keeping the diffusion prior fixed, our method achieves robust results while being trained on a fraction of synthetically labelled data compared to competing data-driven methods, like Depth Anything~\cite{Yang2024_DepthAnything}. The competitive performance of PrimeDepth advocates for the richness of the preimage of large-scale generative models as a starting point for downstream tasks, potentially also beyond depth estimation. One avenue of future research is to explore the complementary nature of data-driven and diffusion-based approaches, since the latter may generalise better, see \cite{DMP_Lee_2024_CVPR} and \cref{fig:painted_image} as an example. We sketch in supplement how a preimage can be integrated into a DPT-based~\cite{Ranftl2021DPT} approach, as used in Depth Anything. 

\begin{figure}[tb]
  \centering
  \includegraphics[width=.98\linewidth]{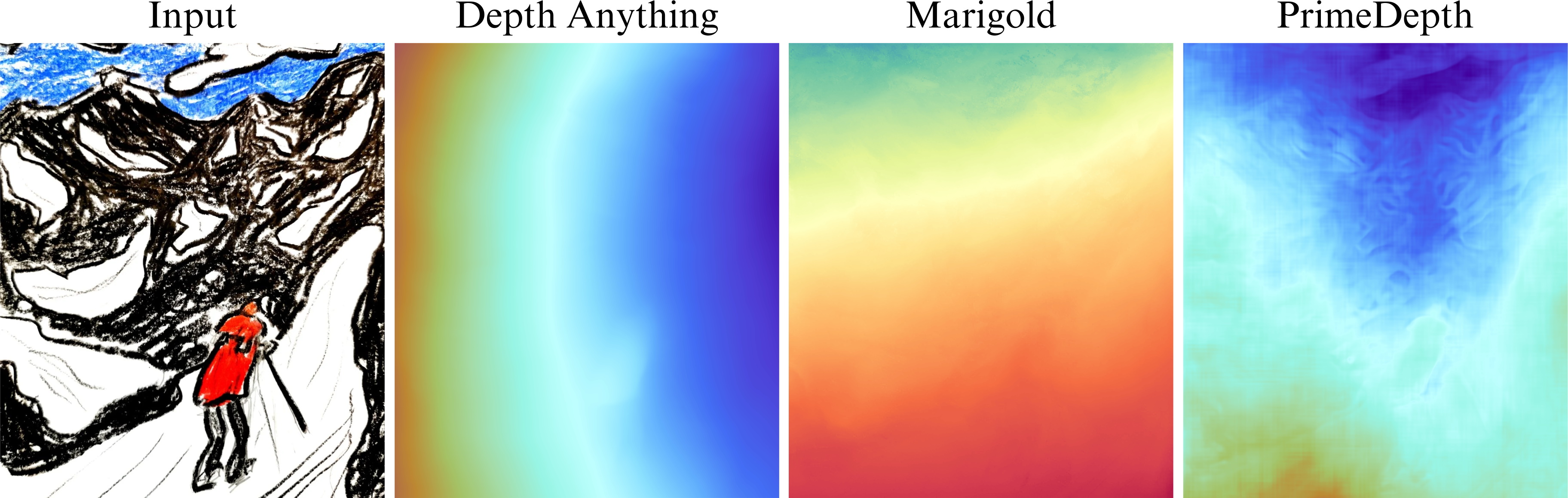}
  \caption{A wax crayon drawing of a mountaineer in rocky and snowy mountains. All methods perform very differently on such an, arguably, ambiguous scene. However, in contrast to Marigold~\cite{Marigold_Ke_2024_CVPR} and our PrimeDepth, Depth Anything~\cite{Yang2024_DepthAnything} fails to capture the general perspective of the scene. This indicates a better generalization capability of diffusion-based methods, such as Marigold and PrimeDepth.}
  
  \label{fig:painted_image}
\end{figure}

\section*{Acknowledgements}
We thank Yannick Pauler his help with evaluation and the enriching discussions. We thank Nicolas Bender for his help in conducting experiments and Friedrich Feiden for his help with illustrations. The project has been supported by the Konrad Zuse School of Excellence in Learning and Intelligent Systems (ELIZA) funded by the German Academic Exchange Service (DAAD). 
The authors gratefully acknowledge the support by the Ministry of Science, Research and the Arts Baden-Württemberg (MWK) through bwHPC, SDS@hd and the German Research  Foundation (DFG) through the grants INST 35/1597-1 FUGG and INST 35/1503-1 FUGG.

%
%
\bibliographystyle{splncs04}
\bibliography{main, accv}

\appendix

\section*{Appendix}

\section{Implementation Details}
\label{supsec:implementation_details}

All datasets were captionised with BLIP-2~\cite{BLIP2_Krause_2023_PLMR}. For pre-training on unlabelled data with a subset of LAION-Aesthetics~\cite{Schuhmann2022_LaionAE}, we predicted the pseudo ground truth for depth with Depth Anything~\cite{Yang2024_DepthAnything} and for segmentation maps with InternImage-H~\cite{wang2022internimage}. We used a training size of resolution $512\times512$ and train with a batch size of 16 for 75K iterations and a learning rate of $10^{-5}$. For the training with labelled data, we followed the protocol of Marigold~\cite{Marigold_Ke_2024_CVPR} and use a combination of Hypersim~\cite{roberts_Hypersim} and Virtual KITTI 2~\cite{cabon2020vkitti2} with a ratio of 9:1. We reduced the learning rate to $5\times10^{-6}$ and train with a batch size of 16 for further 5K iterations.

\section{Inference Speed Comparison}
\label{supsec:samplespeed}
In \cref{suptab:inference_speed}, we compare the inference speed of our model to Depth Anything~\cite{Yang2024_DepthAnything} and Marigold~\cite{Marigold_Ke_2024_CVPR}. For Marigold, we use the same hyper-parameters as specified in the paper, i.e. 50 steps with an ensemble size of 10. 
We see that our method is on average over 120 times faster than the other diffusion-based approach, Marigold, since we do a single denoising step.
Compared to Depth Anything, our method remains around 4 times slower while Marigold is on average 540 times slower. Hereby, we considerably reduce the gap between diffusion-based and transformer-based approaches.
\begin{table}[tb]
  \caption{\textbf{Inference Speed Comparison.} Comparison is performed on an A100 GPU over 1000 samples and reported in seconds. Numbers are the average time it takes to estimate the depth for one image.}
  \label{suptab:inference_speed}
  \centering
  \resizebox{\textwidth}{!}{
  \begin{tabular}{cccccc} 
    \toprule
    Resolution \qquad\qquad
    & PrimeDepth\qquad\qquad
    & Marigold~\cite{Marigold_Ke_2024_CVPR}\qquad\qquad
    & Factor Ours
    & Depth 
    & Factor Ours
    \\
     \qquad\qquad
    & (Ours)\qquad\qquad
    & \qquad\qquad
    & is faster
    & Anything~\cite{Yang2024_DepthAnything}
    & is slower
    \\
    \midrule
    $256\times256$
                    &   0.15    
                    &   19.26    
                    &   128.4   
                    &    0.03   
                    &   5.0       
    \\
    $512\times512$
                    &    0.39   
                    &    43.16  
                    & 110.7     
                    &    0.08   
                    & 4.9         
    \\
    $768\times768$
                    &   0.83    
                    &   105.61   
                    & 127.2         
                    &     0.22   
                    & 3.8     
    \\
    $1024\times1024$
                    &    1.72   
                    &    221.79  
                    & 128.9         
                    &    0.44   
                    & 3.9         
    \\
  \bottomrule
  \end{tabular}
  }
\end{table}

\section{Details for NuScenes-C Dataset}
\label{supsec:dataset_details}

\begin{figure}[tb]
  \centering
  \includegraphics[width=.98\linewidth]{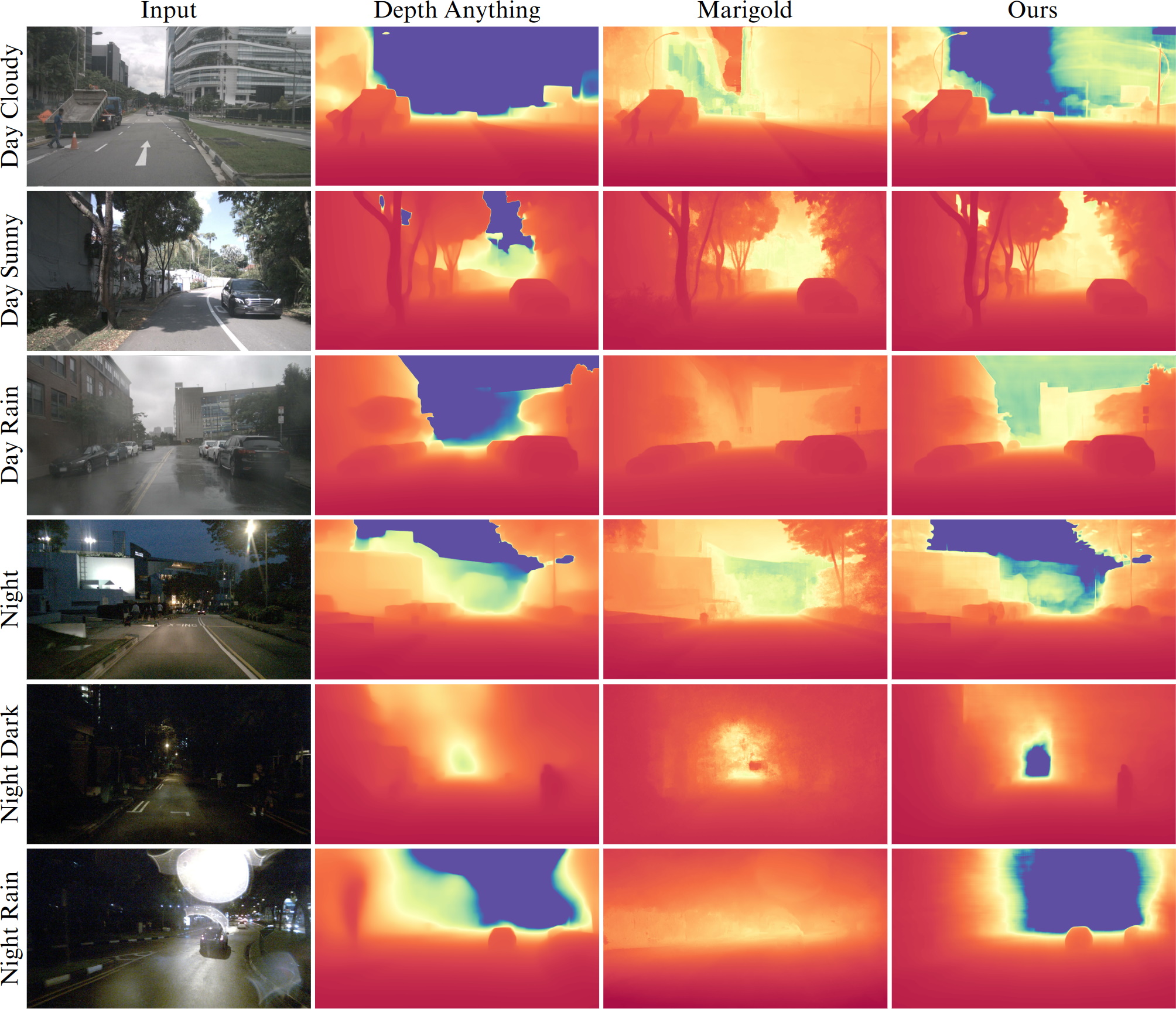}
  \caption{Image sample for each category in nuScenes-C. ``Day Cloudy`` presents a harder challenge for sky separation, ``Day Sunny`` includes sharper shadows, ``Day Rain`` exhibits rain drops on the lens and water reflections, ``Night`` contains multiple light sources and less global illumination, ``Night Dark`` has further reduced visual cues, and ``Night Rain``, compared to ``Day Rain``, additionally contains light refraction artefacts due to water drops.}
  \label{supfig:nuscenes_category_samples}
\end{figure}

The selected scenes of our manually curated dataset of challenging scenarios are shown in \cref{suptab:nuscenes_split}. We split the scenes into the corresponding category based on the text description available in nuScenes ~\cite{caesar2020nuscenes} and visual validation. In total nuScenes-C consists of 3489 images. One image for each category is shown in \cref{supfig:nuscenes_category_samples}.

\begin{table}[tb]
  \caption{The corresponding scenes of each category in our nuScenes-C dataset, sourced from nuScenes ~\cite{caesar2020nuscenes}. Scene identifier denotes the scene name from the original database, formatted as scene-\textit{xxxx}.}
  \label{suptab:nuscenes_split}
  \centering
  \resizebox{\textwidth}{!}{
  \begin{tabular}{llcc} 
    \toprule
    Category &
    Scene identifiers &
    Num. &
    Num. 
    \\
    &
    & Scenes
    & Images
    \\
    \midrule
                    \multirow{2}{*}{Day Cloudy}
                    & 0062, 0064, 0094, 0098, 0124, 0133, 0151, 0159, 0213, 0245,
                    & \multirow{2}{*}{20}
                    & \multirow{2}{*}{795}
                    \\
                    & 0250, 0268, 0347, 0348, 0350, 0355, 0421, 0423, 0740, 0856
                    & 
                    \\
                    \midrule
                    
                    \multirow{2}{*}{Day Sunny}        
                    & 0039, 0041, 0042, 0048, 0052, 0061, 0067, 0162, 0293, 0318,
                    & \multirow{2}{*}{18}
                    & \multirow{2}{*}{722}
                    \\
                    & 0394, 0398, 0416, 0505, 0563, 0791, 0792, 0961
                    &
                    &
                    \\
                    \midrule

                    Day Rain
                    & 0441, 0443, 0452, 0463, 0477, 0573, 0808, 0902
                    & 8
                    & 323
                    \\
                    \midrule

                    \multirow{2}{*}{Night}
                    & 0999, 1001, 1002, 1004, 1005, 1008, 1009, 1015, 1016, 1017,
                    & \multirow{2}{*}{19}
                    & \multirow{2}{*}{765}
                    \\
                    & 1018, 1020, 1049, 1050, 1075, 1077, 1078, 1079, 1083
                    &
                    &
                    \\
                    \midrule

                    Night Dark
                    & 1022, 1023, 1024, 1058, 1060, 1086, 1087, 1091
                    & 8
                    & 321
                    \\
                    \midrule
                    
                    \multirow{2}{*}{Night Rain}
                    & 1048, 1057, 1066, 1070, 1074, 1081, 1094, 1101, 1102, 1104,
                    & \multirow{2}{*}{14}
                    & \multirow{2}{*}{563}
                    \\
                    & 1106, 1107, 1109, 1110
                    &
                    &
                    \\
  \bottomrule
  \end{tabular}
  }
\end{table}

\section{Preimage Block Integration}
\label{supsec:VPD_comparison}
\begin{figure}[tb]
      \centering
      \begin{subfigure}{0.48\linewidth}
      \centering
          \includegraphics[width=0.8\linewidth]{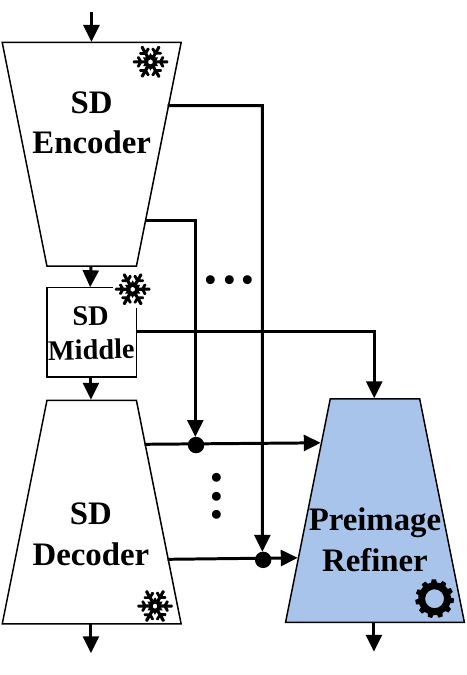}
          \caption{Ours}
      \label{supfig:primrose_architecture}
      \end{subfigure}
      \begin{subfigure}{0.48\linewidth}
      \centering
          \includegraphics[width=0.8\linewidth]{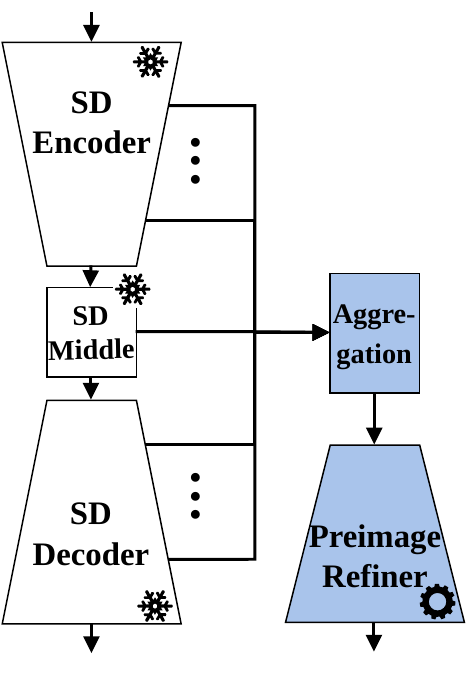}
          \caption{VPD-like}
      \label{supfig:VPD_like_architecture}
      \end{subfigure}

      \caption{
      Architecture comparison of our PrimeDepth (left) and VPD~\cite{VPD_Zhao_2023_ICCV}-like block integration (right). PrimeDepth uses a gradual integration at the corresponding stages while VPD-like architectures use a block-wise integration of the whole preimage as direct input.
      }
      \label{supfig:VPD_architecture}
    \end{figure}

\begin{figure}[htb!]
      \centering
      \includegraphics[width=.98\linewidth]{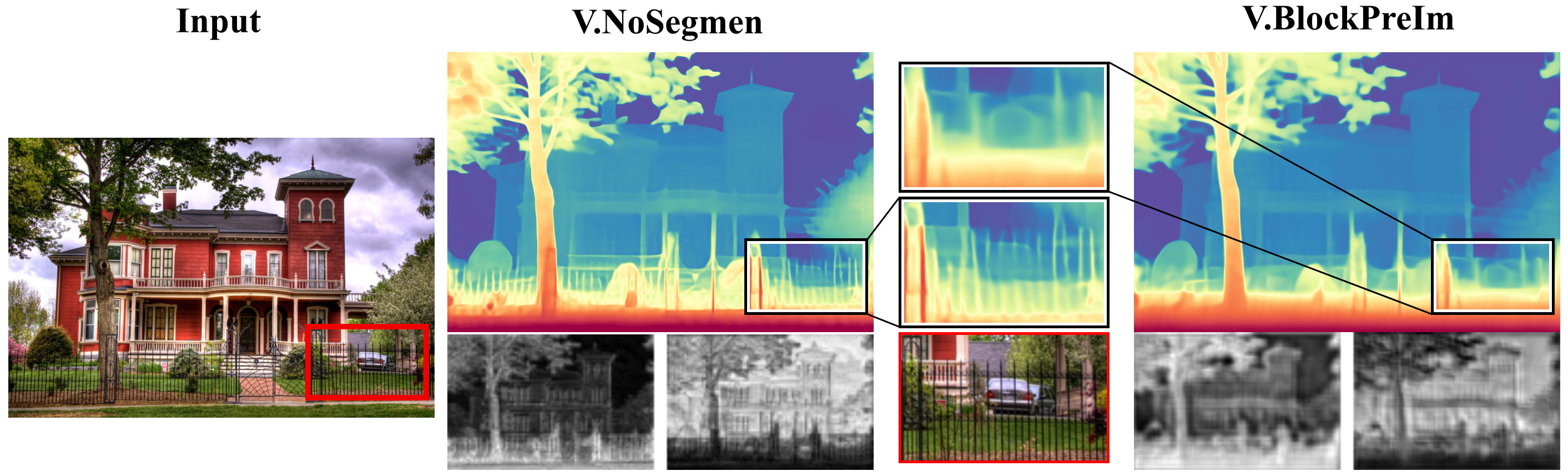}
      \caption{
      Comparison of details from our preimage aggregation (middle) with the simple block aggregation (right) used by methods like VPD~\cite{VPD_Zhao_2023_ICCV}. Both, the depth maps (upper row) and the self-attention maps (lower row) clearly exhibit fewer details when the block aggregation is used. 
      }
      \label{supfig:vdp_qualitative}
    \end{figure}

\cref{supfig:VPD_architecture} illustrates our PrimeDepth architecture (\cref{supfig:primrose_architecture}) in comparison to an alternative architecture which uses a block aggregation (\cref{supfig:VPD_like_architecture}), and we called it V.BlockPreIm in the ablation study. Note that the V.BlockPreIm architecture has been used in previous works like VPD~\cite{VPD_Zhao_2023_ICCV}. The key difference to our architecture is that we use an inductive bias given by the preimage. In our design, the preimage is processed at different stages, i.e. resolutions, of the preimage refiner. In contrast to this, the block aggregation transforms all preimage components into one single block. Because of the tremendous size of the SD preimage, such block approaches have two main disadvantages. Firstly, the aggregated features are highly reduced in channel dimension to be processable by the downstream network. Secondly, features from different resolutions have to be aligned. Hence, one has to decide on a resolution, which is usually smaller than the maximal feature resolution due to feasibility reasons. Features with smaller resolutions and features with larger resolutions therefore have to be up-scaled or down-sized, respectively, which introduces potential redundancies or a loss of information for fine details. We follow the implementation of VPD and and scale the block to the resolution of the second, out of four, stages. 
Our method does not have these drawbacks. In \cref{supfig:vdp_qualitative}, we show an example for the prediction of these two methods (see Sec. 4.4) after complete training. We can see that the block method V.BlockPreIm does indeed fail to predict fine details in comparison to our design (V.NoSegmen). This tendency is also visible in the last self-attention maps of the preimage refining decoder of both methods (\cref{supfig:vdp_qualitative} bottom row). The self-attention maps show a high amount of detail for our method, compared to the blurry ones in the block utilisation.

\section{A Sketch for Integration into DPT}
\label{supsec:dpt_integration}

Our preimage is derived from Stable Diffusion~\cite{LDM_Rombach_2022_CVPR}, and it can be seen as a deconstruction of an image into certain features. Therefore it is an alternative, multi-resolution image representation that can be used instead of the image. A key architectural design of our PrimeDepth has been to use a preimage refiner architecture with an inductive bias, i.e. the different preimage parts are integrate at the respective preimage refiner resolution. When integrating a preimage into other architectures, such as DPT, we believe that it is important that the architectural inductive bias is kept.   

\begin{figure}[tb]
      \centering
      \begin{subfigure}{0.99\linewidth}
      \centering
          \includegraphics[width=0.99\linewidth]{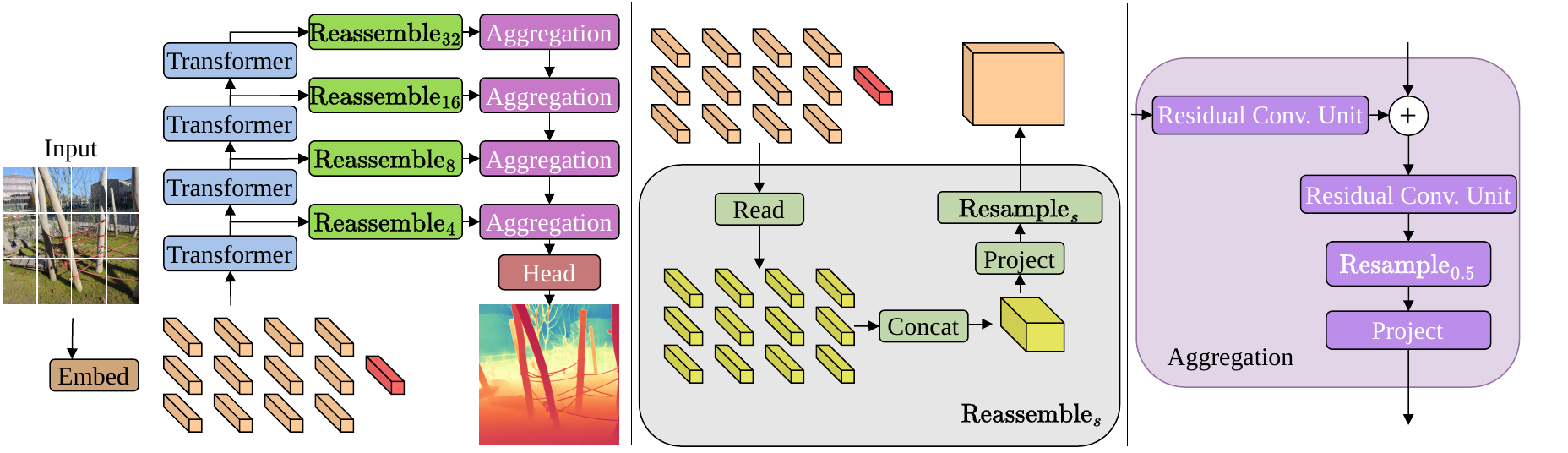}
          \caption{Original DPT~\cite{Ranftl2021DPT} pipeline. Please see \cite{Ranftl2021DPT} for details.\vspace{1cm}}
      \label{supfig:DPT_architecture}
      \end{subfigure}
      \begin{subfigure}{0.99\linewidth}
      \centering
          \includegraphics[width=0.99\linewidth]{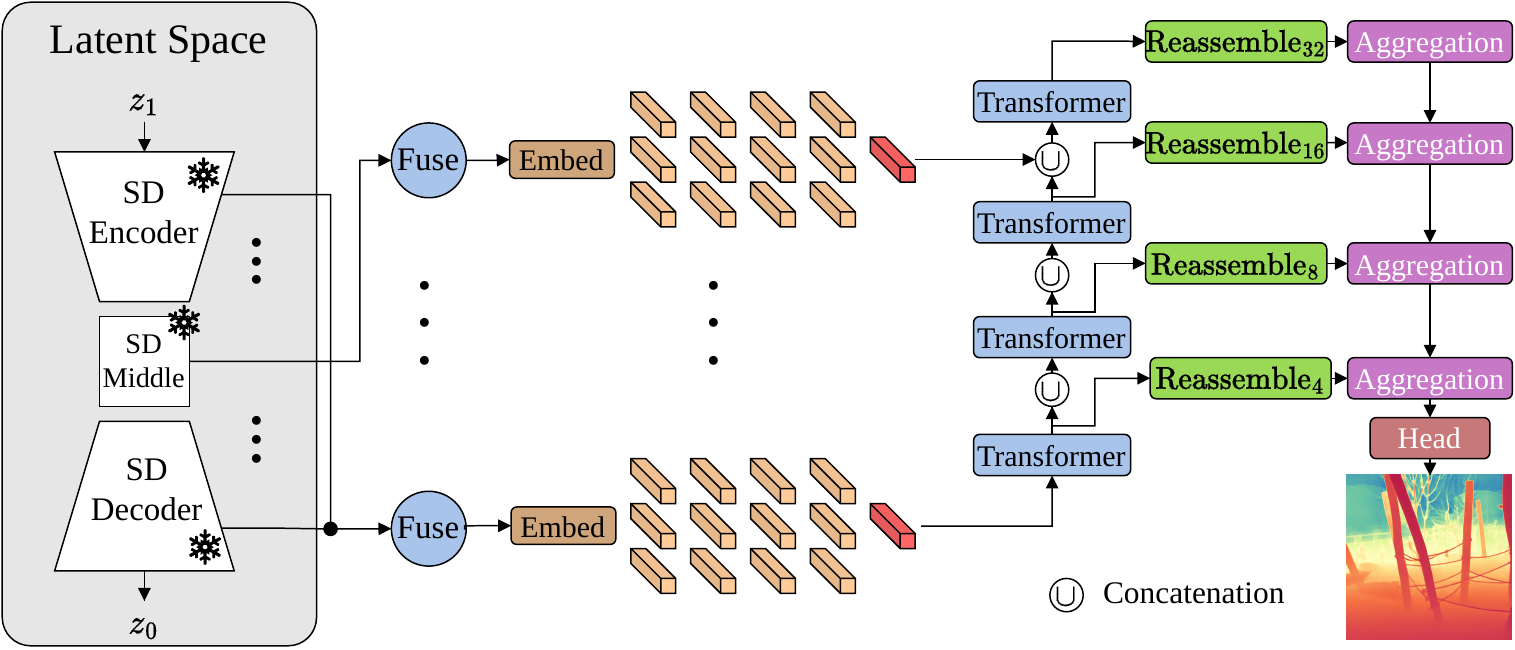}
          \caption{A Sketch for how a preimage may be integrated into the DPT~\cite{Ranftl2021DPT} pipeline. The preimage stages are divided into non-overlapping patches with a positional encoding and a readout token (red block) and provided to the transformer sequence via concatenation.}
      \label{supfig:DPT_integration}
      \end{subfigure}
      \caption{A Sketch for how a preimage may be integrated into the DPT~\cite{Ranftl2021DPT} pipeline}
      \label{supfig:DPT_integration_whole}
    \end{figure}

In \cref{supfig:DPT_integration_whole} we sketch a possible integration into the DPT~\cite{Ranftl2021DPT} model that uses the Vision Transformer~\cite{ViT_Dosovitskiy_2021_ICLR} for depth estimation. \cref{supfig:DPT_architecture} is taken from the corresponding paper~\cite{Ranftl2021DPT} and we refer the reader to the original text for further details. In short, the input image is divided into non-overlapping patches that are adapted with a positional encoding and a readout token (red block)
and fed into transformer blocks. The outputs from the transformer stages are reassembled into image-like representations with different resolutions. They are then processed by a convolutional decoder with gradual integration of different resolution stages. 

A potential integration of the preimage of Stable Diffusion into such a network is sketched in \cref{supfig:DPT_integration}. It follows the principle of the inductive bias by dividing the fused preimage parts into non-overlapping patches and concatenating them into the corresponding resolution-stages processed by the sequence of transformer blocks.

\section{Additional Qualitative Examples}
\label{supsec:qualitative}




    \subsection{Comparison to other methods}

\begin{figure}[tb]
      \centering
      \includegraphics[width=.98\linewidth]{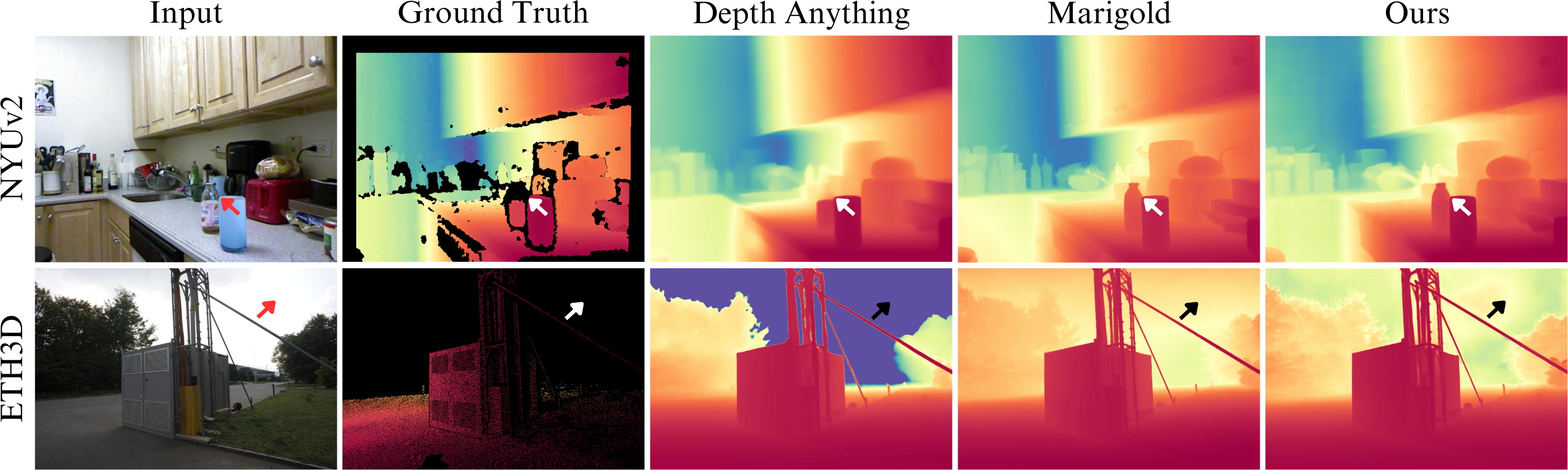}
      \caption{Two failure cases. Depth Anything sees through transparent surfaces, for example a bottle (top row). Both Marigold and our method sometimes struggle with predicting the sky region at infinity (bottom row).}
      \label{supfig:failure_cases}
    \end{figure}

In Fig 4 of the main article, we referred to a failure case of Depth Anything, as well as Marigold and Our PrimeDepth, which we illustrate in  \cref{supfig:failure_cases}. Depth Anything sees through transparent surfaces, for example a bottle (top row). Both, Marigold and our method sometimes struggle with predicting the sky region at infinity (bottom row). 

    \subsection{Comparison for Images in-the-wild}
    We present additional qualitative results for images in-the-wild for our method, Marigold~\cite{Marigold_Ke_2024_CVPR} and Depth Anything~\cite{Yang2024_DepthAnything}. Marigold predicts many details, but at the cost of occasional grainy outputs. It also often struggles with the prediction of sky regions and does not generalise well to unusual images, like a case with watermarks in the image (see image with the palm tree and the sunset in \cref{supfig:inthewild_set2}). Depth Anything appears to yield overall good predictions, but in general does not perform well for detailed depth. Our method predicts detailed depths, as well as also generalising well for difficult images. 

        \begin{figure}[tb]
          \centering
          \includegraphics[width=.98\linewidth]{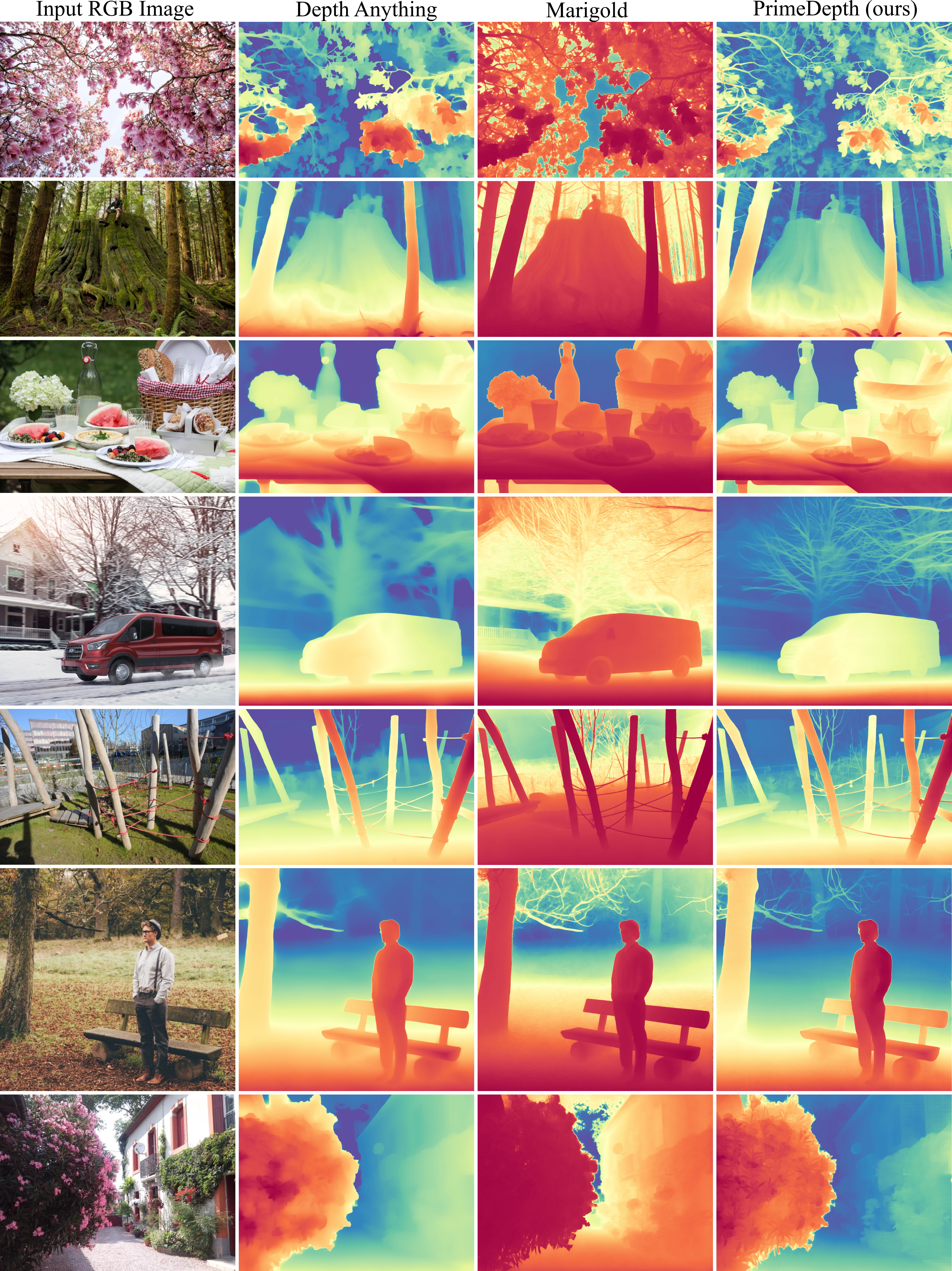}
          \caption{Qualitative comparison of our PrimeDepth to state-of-the-art methods Marigold~\cite{Marigold_Ke_2024_CVPR} and Depth Anything~\cite{Yang2024_DepthAnything} for images in-the-wild. Marigold struggles with sky regions, while Depth Anything provides blurry estimates. In contrast, our model robustly estimates detailed depths.}
          \label{supfig:inthewild_set2}
        \end{figure}

        \begin{figure}[tb]
          \centering
          \includegraphics[width=.98\linewidth]{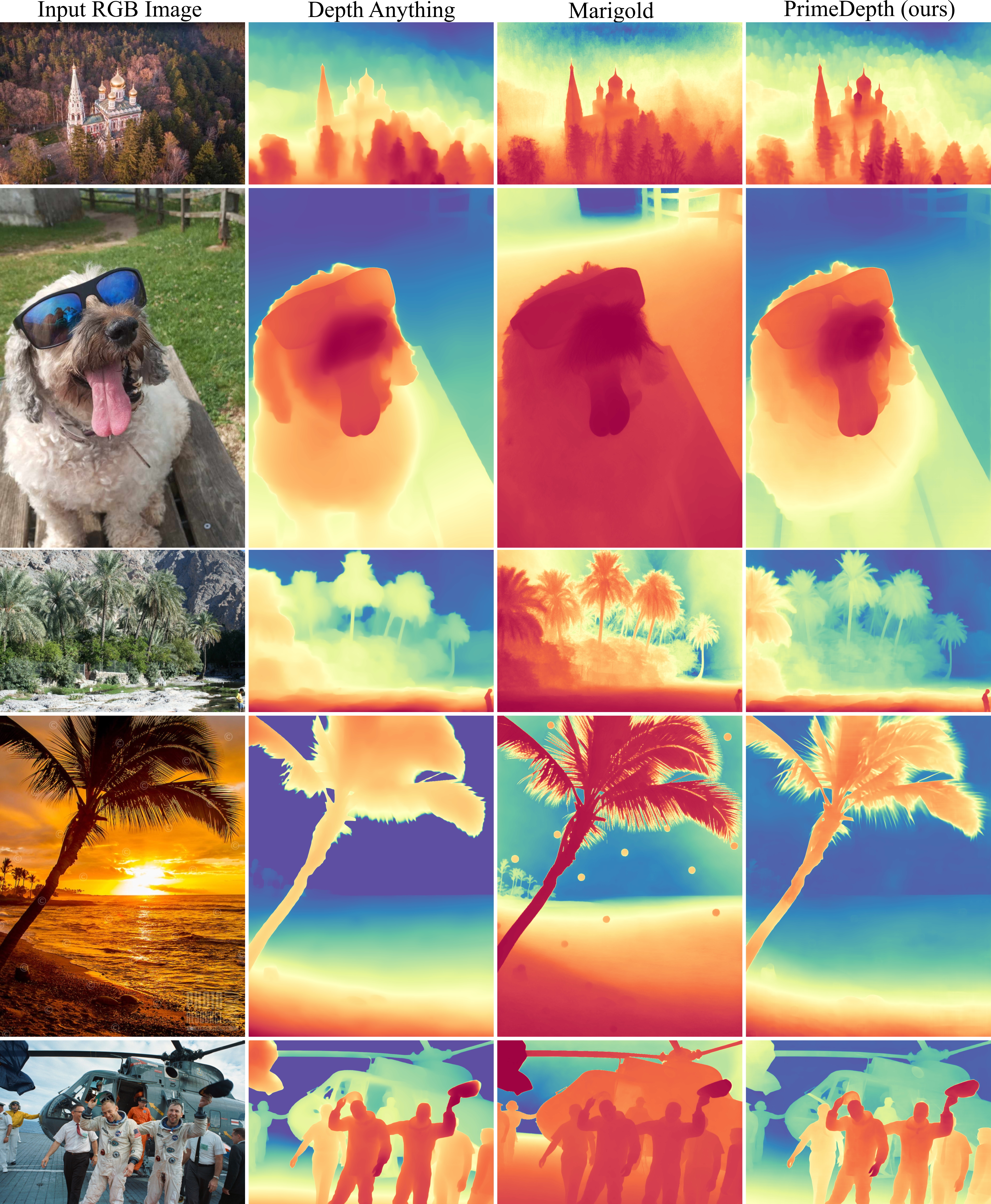}
          \caption{Qualitative comparison of our PrimeDepth to state-of-the-art methods Marigold~\cite{Marigold_Ke_2024_CVPR} and Depth Anything~\cite{Yang2024_DepthAnything} for images in-the-wild. Marigold struggles with sky regions, tends to produce grainy estimates and does not generalise to watermarks. Depth Anything provides robust but less detailed estimates. In contrast, our model robustly estimates detailed depths.}
          \label{supfig:inthewild_set2}
        \end{figure}

\end{document}